\documentclass[10pt,twocolumn,letterpaper]{article}

\usepackage{iccv}
\usepackage{times}
\usepackage{epsfig}
\usepackage{graphicx}
\usepackage{amsmath}
\usepackage{amssymb}

\usepackage[caption=false,font=footnotesize]{subfig}
\usepackage{floatrow}
\usepackage{amsfonts}
\usepackage{todonotes}
\usepackage{multirow}
\usepackage{color,soul}
\usepackage{tikz}
\usepackage{xcolor}
\usepackage{intcalc}
\usepackage{ifthen}
\usetikzlibrary{patterns,arrows,calc,shapes}
\usetikzlibrary{decorations.markings,decorations.pathreplacing}

\newcommand{\myarrow}{-latex}

\usepackage[pagebackref=true,breaklinks=true,letterpaper=true,colorlinks,bookmarks=false]{hyperref}

\newcommand*{\ARXIV}{} % *** Uncomment for arxiv paper

\ifdefined\ARXIV
	\iccvfinalcopy % *** Uncomment for final paper
\else
	\ificcvfinal\fi
\fi

\def\iccvPaperID{2} % *** Enter the ICCV Paper ID here

\begin{document}

%%%%%%%%% TITLE
\title{Short-Term Prediction and Multi-Camera Fusion on Semantic Grids}

\author{Lukas Hoyer\\
Bosch Center for Artificial Intelligence\\
Robert-Bosch-Campus 1, Renningen, Germany\\
{\tt\small lukas.hoyer@outlook.com}
\and
Patrick Kesper\\
Bosch Center for Artificial Intelligence\\
Robert-Bosch-Campus 1, Renningen, Germany\\
{\tt\small patrick.kesper@de.bosch.com}
\and
Anna Khoreva\\
Bosch Center for Artificial Intelligence\\
Robert-Bosch-Campus 1, Renningen, Germany\\
{\tt\small anna.khoreva@de.bosch.com}
\and
Volker Fischer\\
Bosch Center for Artificial Intelligence\\
Robert-Bosch-Campus 1, Renningen, Germany\\
{\tt\small volker.fischer@de.bosch.com}
}

\maketitle
\ificcvfinal\thispagestyle{empty}\fi

%%%%%%%%% ABSTRACT
\begin{abstract}
	An environment representation (ER) is a substantial part of every autonomous system.
	It introduces a common interface between perception and other system components, such as decision making, and allows downstream algorithms to deal with abstracted data without knowledge of the used sensor.
	In this work, we propose and evaluate a novel architecture that generates an egocentric, grid-based, predictive, and semantically-interpretable ER.
	In particular, we provide a proof of concept for the spatio-temporal fusion of multiple camera sequences and short-term prediction in such an ER.
	Our design utilizes a strong semantic segmentation network together with depth and egomotion estimates to first extract semantic information from multiple camera streams and then transform these separately into egocentric temporally-aligned bird's-eye view grids.
	A deep encoder-decoder network is trained to fuse a stack of these grids into a unified semantic grid representation and to predict the dynamics of its surrounding.
	We evaluate this representation on real-world sequences of the Cityscapes dataset and show that our architecture
	can make accurate predictions in complex sensor fusion scenarios and significantly outperforms a model-driven baseline in a category-based evaluation.
\end{abstract}

%%%%%%%%% INTRODUCTION
\section{Introduction}\label{sec:introduction}

\begin{figure}
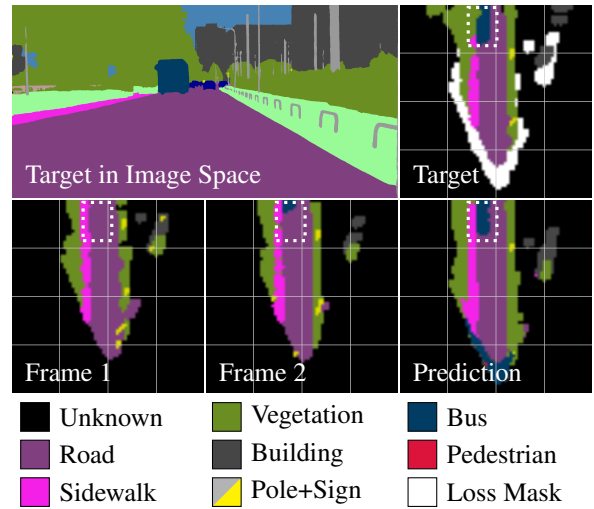

\input{images/catchy_example_t.tex}\\
\input{images/catchy_example_p.tex}%
\caption{(best viewed in color) Semantic grid prediction: Our framework is based on semantic segmentations in image space (top left). These are transformed into the proposed bird's eye view \textit{semantic grid} representation (top right). We feed our architecture with a sequence of input frames (bottom left and bottom center) to predict the subsequent frame (bottom right), which is compared with the target frame (top right). In the shown example, a bus (dark turquoise) is entering the grid at the top (bottom row). One sees that our architecture is able to predict its movement further into the grid even though the vehicle became just visible in frame 2. For details on the loss mask and prediction artifacts, see Sec. \ref{sec:experiments}.}
\label{fig:catchy_example}
\end{figure}

In recent years, deep learning methods have been investigated to control autonomous systems, such as self-driving cars or robots.
An important property of such systems is their capability to perceive complex situations using multiple sensors and to act accordingly in a fast and reliable way. To enable intelligent decision making, a common environment representation (ER) as interface between different sensors and the downstream control has to be provided.

Such an ER should have certain properties. First, it should unify different sensor representations to support sensor fusion. In that way, downstream algorithms can work with this abstract representation without knowledge of the used sensors. Second, the ER should be interpretable to modularize the system and enhance human accessibility. So, the system can be debugged and understood more easily, improving its reliability. And third, it should be predictive to compensate for system-inherent latencies caused by sensor measurements or signal processing, which is particularly important in the context of recent computationally expensive computer vision algorithms.

Due to these reasons, we present and evaluate a proof of concept for generating an egocentric, interpretable, predictive, and grid-based ER, most importantly including the spatio-temporal fusion of semantic information of multiple cameras and short-term prediction. We call this ER \textit{semantic grid}. It is a bird's eye view $2\text{D}$ projection of semantic features of the environment (see Fig. \ref{fig:catchy_example} top row).

\begin{figure}
\input{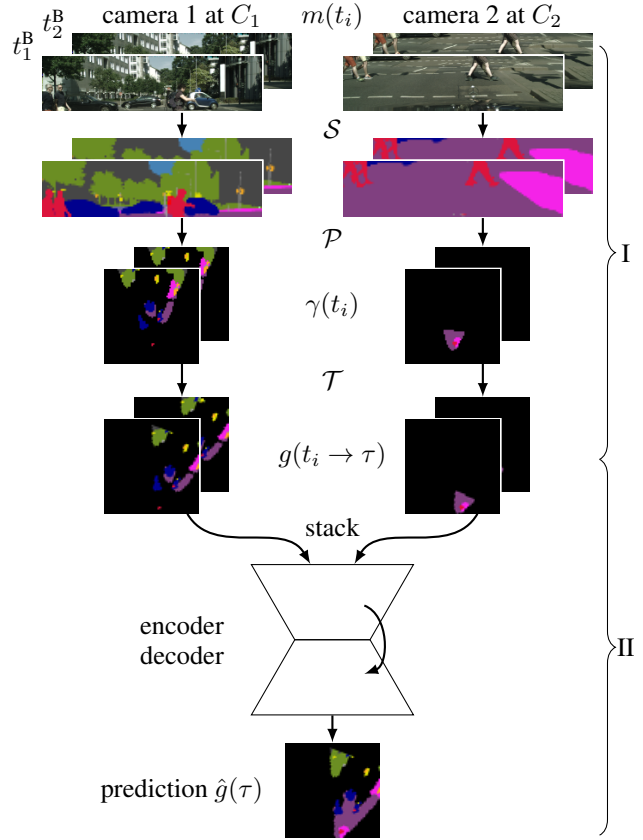}
\caption{(best viewed in color) Schematic overview of our semantic grid fusion and prediction architecture. The framework is fed with image sequences \(m({t_i})\) of different cameras from past times \(t_i\). These are semantically segmented ($\mathcal{S}$), projected ($\mathcal{P}$) into a agent-centric top-down view $2\text{D}$ point cloud \(\gamma({t_i})\) and transformed (\(\mathcal{T}\)) to the same time $\tau$ resulting in spatio-temporally aligned grids $g({t_i} \rightarrow {\tau})$. The encoder-decoder deep neural network (ED) fuses these grids and predicts environment dynamics. The two-step description in the Methods Section is indicated as I and II on the right. Note that the point cloud \(\gamma({t_i})\) is represented as grid for visualization purpose.}
\label{fig:information_flow}
\end{figure}

Our concrete architecture for producing semantic grids can be disentangled into two major parts (see I and II in Fig. \ref{fig:information_flow}):
First, semantic information is extracted from each camera signal using deep neural networks for semantic segmentation (see $\mathcal{S}$ in Fig. \ref{fig:information_flow}).
The semantic information is then spatially transformed into a top-down, bird's eye view semantic grid, using depth information provided by the stereo camera (see $\mathcal{P}$ in Fig \ref{fig:information_flow}).
Grids from different cameras and past time frames are temporally aligned to a certain future time \(\tau\) using the agent's egomotion (see $\mathcal{T}$ in Fig. \ref{fig:information_flow}).
Note that in particular independently moving objects will be transformed incorrectly, as there is no motion model of other objects than the agent itself, yet.

Second, these spatio-temporally aligned representations are combined into a single grid using a deep encoder-decoder (ED) neural network.
This network contains all trainable parameters of our architecture and has the non-trivial task to fuse the grids from different cameras and past times, making assumptions about the environment, as well as predicting the motion of potentially multiple dynamic objects. Such a prediction is shown in Fig. \ref{fig:catchy_example}. Even though in this work we focus on multi-camera fusion, our architecture can be extended to other modalities such as LiDAR given an appropriate algorithm for extracting spatial semantic features from the sensor.

We have evaluated our architecture with respect to different semantic categories on the camera-based Cityscapes dataset of driving scenarios. Our approach significantly outperforms solely model-driven baselines for single camera and multi-camera prediction, considering missing egomotion information, different sequence length, and varying prediction horizons.

To the best of our knowledge, we are the first to investigate a deep convolutional neural network for \textit{short-term prediction} and \textit{multi-camera} sequence fusion on \textit{semantic grids}. We evaluate the proposed architecture on real-world data. Our approach enables autonomous systems to perceive the environment with multiple cameras and helps to mitigate the problem of long runtimes of recent computer vision algorithms.

After giving an overview of related work in Section \ref{sec:related_work} and a general description of the presented approach in Section \ref{sec:methods},
we lay out experimental setups in Section \ref{sec:experiment_setup}, present the results in Section \ref{sec:experiments}, followed by a discussion and conclusion in Sections \ref{sec:conclusion} and \ref{sec:realconclusion} respectively.

%%%%%%%%% RELATED WORK
\section{Related Work}\label{sec:related_work}

Due to the recent discovery of certain cells dedicated to spatial referencing in brains
\cite{hafting2005microstructure}, grid-based representations seem biologically plausible.
One well known example of interpretable grid-based ER are single- and multi-sensor occupancy grids (e.g., \cite{colleens2007occupancy, moravec1989sensor}), which have also gained interest in recent deep learning studies (e.g., \cite{dequaire2018deep, gupta2017cognitive, lundell2018deep}).
Due to the lack of suitable feature extractors before the raise of deep learning, these grids normally only consisted of one semantic feature encoding the occupancy property. However, now semantic grid representations have become a recent subject of interest. For instance, \cite{erkent2018semantic2, erkent2018semantic, lu2018monocular} deal with the generation of single-camera, non-predictive semantic grids, while \cite{bansal2018chauffeurnet} utilize them for decision-making in autonomous driving as semantic grids are easier to simulate than raw sensor data. Other approaches generate sparse bird's eye view representations of the environment by detecting objects in image space and transforming them to the top-down view (e.g., \cite{palazzi2017learning, wang2019monocular}).

We want to differentiate and hence disentangle two different but both desired predictive properties of autonomous agents.
First, the agent should be able to reason about future events, e.g., for planning, and in this sense should be able to form mid- or long term predictions. For instance, \mbox{\cite{bansal2018chauffeurnet, lee2017desire, park2018sequence}} perform explicit long-term vehicle trajectory prediction using RNNs.
From this, we differentiate a second form of prediction, intrinsic to perception.
Sensors and their down-stream signal processing (e.g. semantic segmentation) induce a temporal delay between the actual present state of the world and the agent's belief about this state.
Our aim is to design an environment representation that can compensate these short-term system-inherent latencies and synchronize data from different time horizons.

For short-term prediction of camera data, there are approaches predicting extracted image features such as object bounding boxes \cite{bhattacharyya2018long}, semantic segmentation \cite{luc2017predicting}, or instance segmentation \cite{luc2018predicting}. However, all of these work are in the sensor-dependent image space, which unnecessarily complicates the task considering downstream sensor fusion.
In contrast, \cite{dequaire2018deep, dequaire2016deep, ondruska2016end, ondruska2016deep} and \cite{schreiber2018long} employ an end-to-end trainable recurrent architecture to directly predict an unoccluded occupancy grid from laser data, capable to track multiple objects.
We distinguish ourselves by investigating the capacity of our architecture in the context of sensor fusion and further,
by using camera data instead of laser data, enabling semantically richer representations. Moreover, we combine semantic predictions with a moving sensor instead of analyzing both scenarios separately.

% ## Detection in image and transformation to BEV

% Palazzi17 Learning to Map Vehicles into Bird's Eye View
% * dataset based on GTA5 for frontal view to bird's eye view RGB
% * trained network to warp detections from first to second view

% Wang19 Monocular Plan View Networks for Autonomous Driving
% * We detect vehicles and pedestrians in the first person view and project them into an overhead plan view. This representation provides an abstraction of the environment from which a deep network can easily deduce the positions and directions of entities.

% ## 3D Semantic Point Cloud

% Li19 Fast 3D Semantic Mapping in Road Scenes
% * 3D Semantic Mapping with monocular camera
% * for road scenes

% ## Long-term prediction

% ## Multi-view fusion

% Zhu18 Generative Adversarial Frontal View to Bird View Synthesis
% * generate RGB bird's eye view from single frontal view image

% Zhang15 Sensor Fusion for Semantic Segmentation of Urban Scenes
% * fusion of camera and 3d point cloud for semantic segmentation

% Luo18 Fast and Furious: Real Time End-to-End 3D Detection, Tracking and Motion Forecasting with a Single Convolutional Net
% * deep neural network that is able to jointly reason about 3D detection, tracking and motion forecasting given data captured by a 3D sensor
% * no RGB information

% Sobh End-To-End Multi-Modal Sensors Fusion System For Urban Automated Driving
% *

% Shiloh01 Multiple Camera Fusion for Multi-Object Tracking
% *

%%%%%%%%% METHODS
\section{Methods}\label{sec:methods}

\begin{figure*}
\centering
\includegraphics[width=6in]{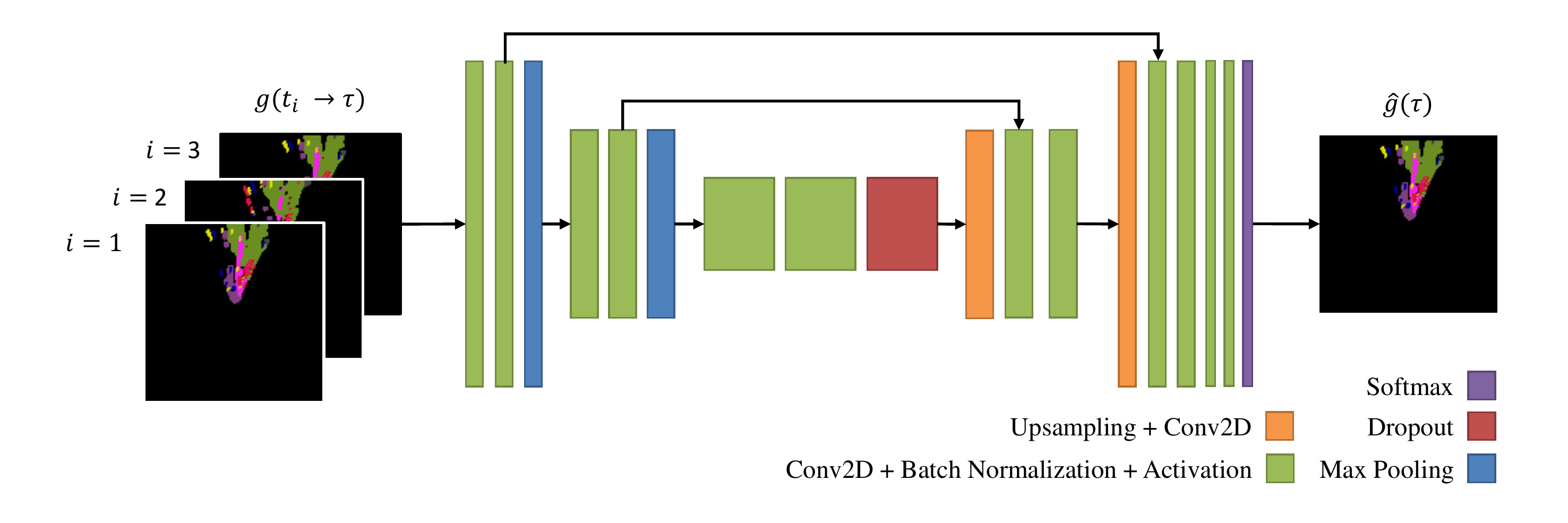}
\caption{(best viewed in color) Encoder-decoder CNN (ED). The ED combines synchronized semantic grids into one prediction including dynamic object transformation, temporal filtering, and multi-camera fusion.}
\label{fig_encoder_decoder}
\end{figure*}

In this section, we describe our architecture that generates semantic grids from semantic segmentations of camera images (see Fig.~\ref{fig:information_flow}). A \textit{semantic grid} is a \(g \in \mathbb{R}^{W_G \times H_G \times F}\),
with \(W_G, H_G, F \in \mathbb{N}\) denoting the grid's width, height, and number of semantic features (e.g. different object classes) respectively.
Each grid cell represents a certain area in space and hence the spatial coordinate is determined by
the cell's indices.
The semantic grid is \textit{egocentric} in the sense that the agent is constantly located at the same grid cell.

\textbf{Generating synchronized semantic grids (Fig. \ref{fig:information_flow}, I):}
Given a RGB image \(m \in \mathbb{R}^{W_I \times H_I \times 3}\) and a depth map \(d \in \mathbb{R}^{W_I \times H_I}\),
where \(W_I, H_I \in \mathbb{N}\) denote the pixel resolution,
we first use some semantic segmentation network \(\mathcal{S}\) to generate semantic information in image space \(s=\mathcal{S}(m) \in [0,1]^{W_I \times H_I \times F}\).

Second, this sensor-dependent representation (in image space) is transformed into a spatially aligned (w.r.t. the agent) continuous \(2\text{D}\) point cloud \(\gamma \in \mathbb{R}^{W_I \times H_I \times (2+F)}\), using the depth map \(d\) to project the semantic information from pixel space to the floor plane of the camera coordinate system at recording time \(t_i\):
% This representation is used to avoid quantization errors during the transformations.

\begin{equation}
	\gamma({t_i}) = \mathcal{P}(s({t_i}), d({t_i}))
	\label{eq:projection}
\end{equation}
For that purpose, each segmentation pixel $s_{uv} \in \mathbb{R}^F$ with $u \in \{1,\dots,W_I)$ and $v \in \{1,\dots,H_I\}$ is transformed to a \(2\text{D}\) point with semantic information
$\gamma_{uv} = (\gamma_{uv}^x, \gamma_{uv}^z, \gamma_{uv}^F)^T \in \mathbb{R}^{2} \times \mathbb{R}^F$
in the agent coordinate system $A$. This transformation is achieved using the intrinsic camera matrix $K$ and the extrinsic parameters $R_{C\rightarrow A}$ and $t_{C\rightarrow A}$ describing the camera pose $C$ with respect to the agent%
\ifdefined\ARXIV
~(see supplementary material for details on those matrices).
\else
~(see \cite{cityscapes_calibration} for details on those matrices).
\fi
\begin{equation}
	\begin{pmatrix}\gamma_{uv}^x \\ \gamma_{uv}^y \\ \gamma_{uv}^z\end{pmatrix} =
	(R_{C \rightarrow A}, t_{C \rightarrow A}) \cdot K^{-1} \cdot d_{uv} \cdot \begin{pmatrix}u \\ v \\ 1\end{pmatrix}
	\label{eq:camera}
\end{equation}
Using these agent-centric \(3\text{D}\) coordinates, the semantic features $s$ are projected onto the ground floor. For simplicity, we define $\mathcal{P}$ as well as the following operators on a single point while they are applied to the whole point cloud.
\begin{equation}
	\mathcal{P}(s_{uv},d_{uv}) := (\gamma_{uv}^x, \gamma_{uv}^z, s_{uv})^T
\end{equation}

Third, the semantic grids are spatially aligned according to the agent's current orientation
\(\alpha_{t_0}^{t_i} = \sum_{j=1}^i \dot{\alpha}_{t_j} \cdot (t_j-t_{j-1})\), which is the integrated angular component of the egomotion, and the point cloud is discretized to the grid representation.

\begin{equation}
	g({t_i}) = \mathcal{D}(\mathcal{T}_1(\gamma({t_i}), \alpha_{t_0}^{t_i}))
\end{equation}

\begin{equation}
	\mathcal{T}_1(\gamma_{uv},\alpha) :=
	\begin{bmatrix}
		\cos{\alpha} & -\sin{\alpha} & \multirow{2}{*}{0}\\
		\sin{\alpha} & \cos{\alpha} & \\
		\multicolumn{2}{c}{0} & \text{Id}_F\\
	\end{bmatrix} \cdot
	\begin{pmatrix}\gamma_{uv}^{x} \\ \gamma_{uv}^z \\ \gamma_{uv}^F\end{pmatrix}
\end{equation}
Thereby, rotations of the agent do not cause rotation of the entire grid, but only result in a change of the agent's internal orientation \(\alpha_{t_0}^{t_i}\) and a rotation of potential new data, reducing quantization errors.
The discretization $\mathcal{D}$ assigns the 2D points $\gamma_{uv}$ to grid cells $g_{kl}$. If two points are allocated to the same grid cell, they are prioritized preferring small and dynamic classes. In that way, we ensure that no important information is occluded by another class. If no point is assigned to a grid cell, it is classified as unknown (black).

And fourth, the integrated translational component of egomotion
\(q_{t_i}^{\tau=t_k} = \sum_{j=i}^{k-1} \dot{q}_{t_j} \cdot (t_{j+1}-t_{j})\)
% \(q_{t_i}^{\tau}\)
is used for the parameter-free temporal extrapolation (grid translation) into the future time \(\tau\):

\begin{equation}
	g({t_i} \rightarrow {\tau}) = \mathcal{T}_2(g({t_i}), q_{t_i}^{\tau}).
\end{equation}

\begin{equation}
	\mathcal{T}_2(g_{kl},q) := g_{k-q^x, l-q^z}
\end{equation}
assuming that $q$ is represented in pixel coordinates. If $k+q^x >= W_G$ or $l+q^z >= H_G$, $g_{kl}$ is assigned the class unknown (black).

Let \(\{m({t_1}), \hdots, m({t_n})\}\) denote an input sequence of length \(n\) of past images at times \(t_1, \hdots, t_n\) from one sensors, then \(\{g({t_1} \rightarrow{\tau}), \hdots, g({t_n} \rightarrow {\tau})\}\) are accordingly synchronized to the same future time \(\tau\) using the parameter-free (and hence not trained) transformations \(\mathcal{P}\) and \(\mathcal{T} = \mathcal{T}_2 \circ \mathcal{D} \circ \mathcal{T}_1\).

\textbf{Predictive fusion architecture (Fig. \ref{fig:information_flow}, II):}
These, now synchronized single-sensor semantic grids, are stacked and fed into an encoder-decoder deep neural network (ED) (see Fig. \ref{fig:information_flow} stack).
Its task is to fuse the grids from multiple sensors and different original time steps and to incorporate object motion.
The ED network yields the predicted semantic grid \(\hat{g}(\tau)\) at time \(\tau\).
In case \(\tau-t_n\) covers the system-inherent latency, \(\hat{g}(\tau)\) represents the agents belief about the actual present situation. The ED is trained using self-supervised learning. It is provided with grid input sequences \(\{g(t_1 \rightarrow \tau), \hdots, g(t_n \rightarrow \tau)\}\) for multiple sensors, while withholding the grid \(g(\tau)\), measured at the time \(\tau > t_n\), as ground truth.

\textbf{Loss and validation metrics:}
For training and validation, we use the categorical cross-entropy loss function and the intersection over union (IoU)~\cite{long2015fully}, respectively. For both, the area $M$ that the sensors have already seen in past frames, but where no ground truth information is available for the target frame, is ignored.

For this purpose, we calculate a masked loss
\begin{equation}
  \mathcal{L}_{\text{masked}} = f(g(\tau), (1-M) \cdot \hat{g}(\tau) + M \cdot g(\tau)),
\end{equation}
where \(g(\tau)\) and \(\hat{g}(\tau)\) are the ground truth and predicted semantic grids respectively, \(f\) is some distance measure (e.g. categorical cross entropy), and \(M\) is the mask of the covered area without ground truth information \(M = M_{\text{covered}} - M_{\text{target}}\).
Here, \(M_{\text{target}}\) denotes the area with ground truth information \(M_{\text{target}} = \text{known}(g(\tau))\) and \(M_{\text{covered}}\) the area the sensors have seen during the entire synchronized sequence (including the target frame \(g(\tau)\)) \(M_{\text{covered}} = \text{known}(g(\tau) + \sum\limits_{i=1}^{n}g(t_i \rightarrow \tau))\). The function "$\text{known}$" determines the area within a grid where the classification is not the class "unknown".
In the area determined through \(M\), the network is not punished for any predictions.

In that way, we want to encourage the network to remember areas, which are newly occluded in the target frame (e.g. the waiting car in Fig. \ref{fig_example_occluded_car} that is newly occluded by another car in the target frame; $M$ visualized in white color),
instead of predicting the class "unknown" (white/black).

%%%%%%%%% EXPERIMENTS
\section{Experiment Setup}\label{sec:experiment_setup}

\textbf{Dataset:} For our experiments, we have chosen the Cityscapes dataset of driving scenarios~\cite{cordts2016cityscapes}, which provides real-world RGB image sequences ($30~\text{frames}$, $17~\text{Hz}$) including the preprocessed disparity of the stereo camera, the egomotion of the vehicle, as well as training data for semantic segmentation. In that way, Cityscapes allows the generation of semantic grids with various and diverse classes.

As some classes are comparatively rare due to their small 2D projection during the semantic grid generation for training, we have combined large vehicles (bus, truck, and train) as well as small static objects (poles, traffic signs, and traffic
lights), which have a similar semantic meaning and behave similarly in the semantic grid representation. Therefore, the color coding for the classes in a semantic grid
\ifdefined\ARXIV
(see Tab. \ref{tab_classresults})
\fi
does not exactly match with the Cityscapes colors.

\begin{figure}
\centering
\vspace{0.2cm}
\input{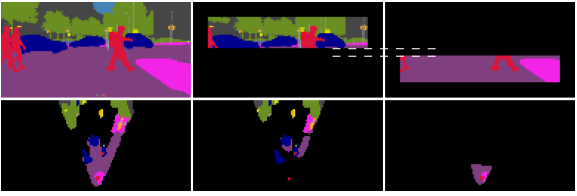}
\caption{(best viewed in color) Sensor Split 2 for simulating multiple sensors. The semantic segmentation of the camera image (top left) is cropped to simulate two separate sensors (top center and right). In the lower row the associated egocentric semantic grids are visualized. The white dashed lines mark the blind spot.}
\label{fig_extended_split}
\end{figure}

To evaluate the performance of the semantic grid prediction with respect to multi-camera fusion, we split the camera frames into separate image sections. While the \textit{Split~1} scenario just divides the images into a lower and an upper part, the \textit{Split~2} scenario additionally has a left and right black margin, as well as a blind area between the lower and upper image section (see Fig.~\ref{fig_extended_split}). These splits are especially challenging due to the vertical split direction, as there are ambiguities when merging the resulting semantic grids. Moreover, the image is split close below the horizon, which is the most crucial part of the image. In that way, we want to underline the capabilities of the ED to solve ambiguities, track objects between both cameras, and conclude information about blind spots.

\textbf{Semantic Grid Generation:}
We generate the semantic segmentation of the Cityscapes sequences using the state-of-the-art deep neural network DeepLabv3~\cite{chen2017rethinking}, which achieves \(80.31\) \% IoU on Cityscapes. The semantic labels are transformed to the semantic grid using the ego motion of the vehicle and the disparity of the stereo camera (see Section \ref{sec:methods}). For the semantic grid, we have chosen a size of \(128 \times 128\) pixels, which is the equivalent of \(100 \times 100\) meters.

\textbf{Dataset of Grid Sequences:}
For training and evaluation of our semantic grid prediction framework, the frames $o, o+s, o+2s, \hdots, o+(n-1)s$, where $o$ is the sequence offset and $s$ the step size, are used as input frames, while frame $o+n\cdot s$ represents the target frame. Here, we have used the step size $s=5$.
The training sequences are overlappingly sampled ($o$ is chosen arbitrarily), whereas the validation sequences are disjoint and their ground truth frames are aligned for the different experiments. In case of $n=2$, we work with 59500 training sequences and 500 validation sequences. All results are reported on the validation dataset.

\textbf{Encoder-Decoder CNN:}
In this paper, we have used an encoder-decoder Convolutional Neural Network (ED CNN) similar to U-Net~\cite{ronneberger2015u}~(see Fig. \ref{fig_encoder_decoder}).
The encoder with depth $d$ contains $d$ down sampling blocks. Each of them reduces the spatial resolution by half and doubles the initial number of features $f$. The decoder consists of $d-1$ up sampling blocks and finally a softmax layer. Between each corresponding down- and upsampling block are skip connections to enable dense predictions \cite{long2015fully}.
\ifdefined\ARXIV
Further information on the implementation and training can be found in the supplementary material.
\fi

\section{Experiments}\label{sec:experiments}

\begin{figure*}
  \centering
  \begin{minipage}{0.49\textwidth}
  \centering
  \subfloat{%
    \label{fig_results_static}%
    \includegraphics[width=3.2in]{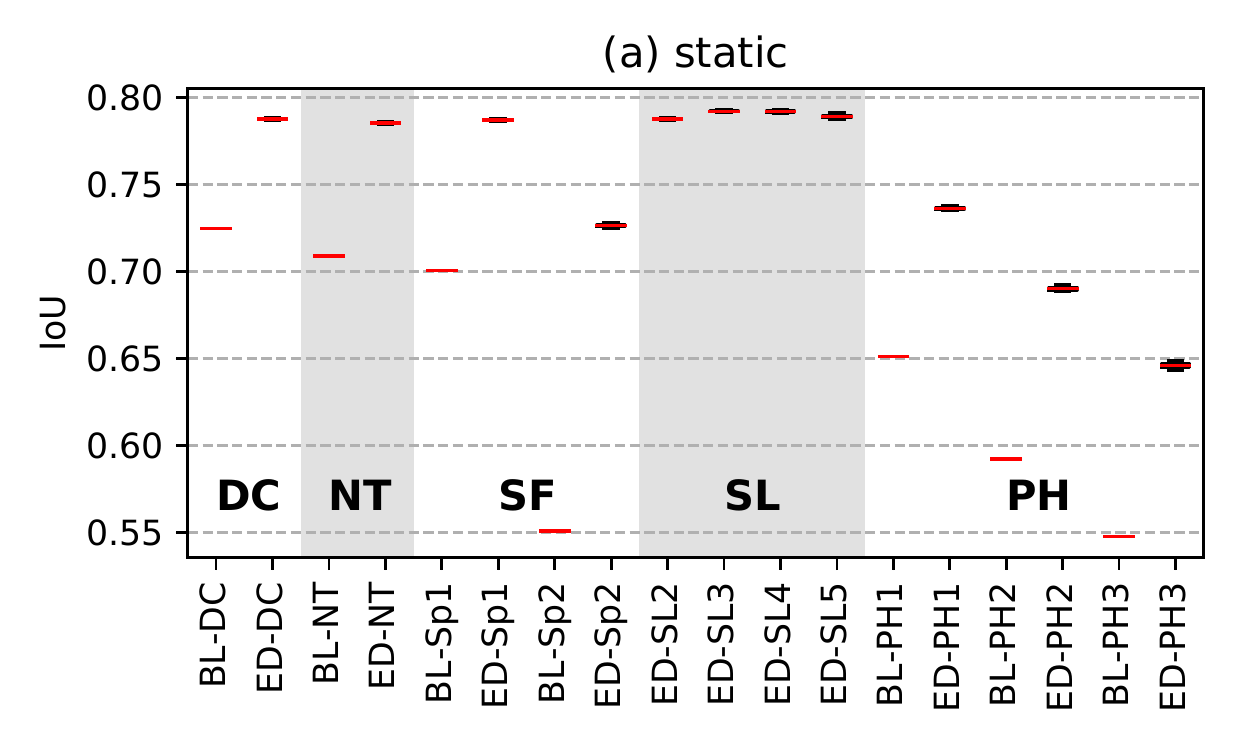}}\\
  \vspace{-0.5cm}
  \centering
  \subfloat{%
    \label{fig_results_small_static}%
    \includegraphics[width=3.2in]{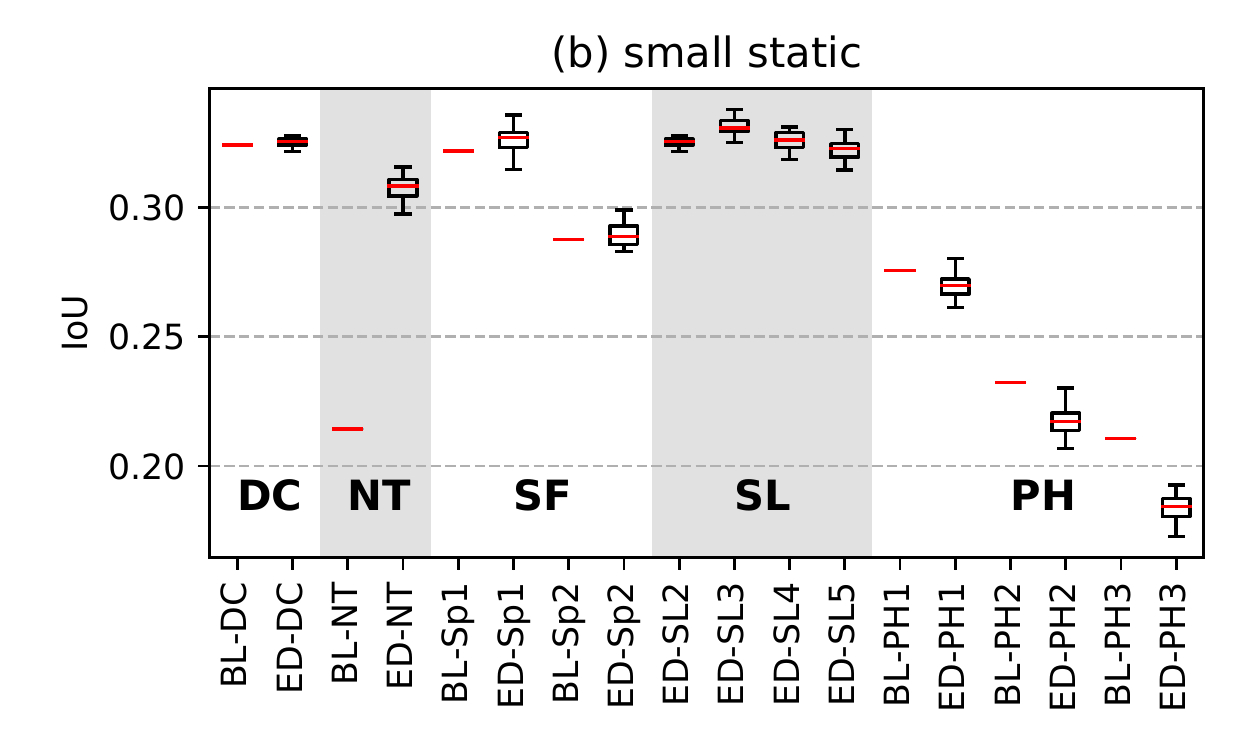}}
  \end{minipage}
  \centering
  \begin{minipage}{0.49\textwidth}
  \centering
  \subfloat{%
    \label{fig_results_vehicles}%
    \includegraphics[width=3.2in]{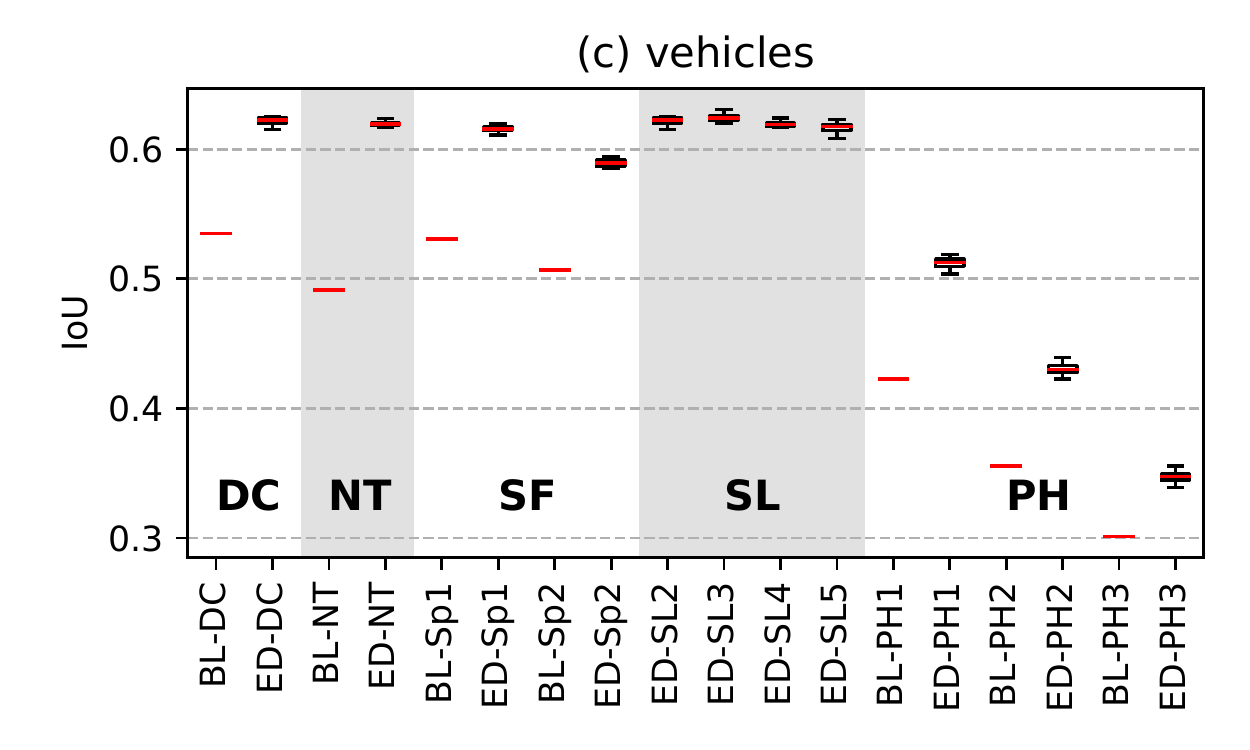}}\\
  \vspace{-0.5cm}
  \centering
  \subfloat{%
    \label{fig_results_small_dynamic}%
    \includegraphics[width=3.2in]{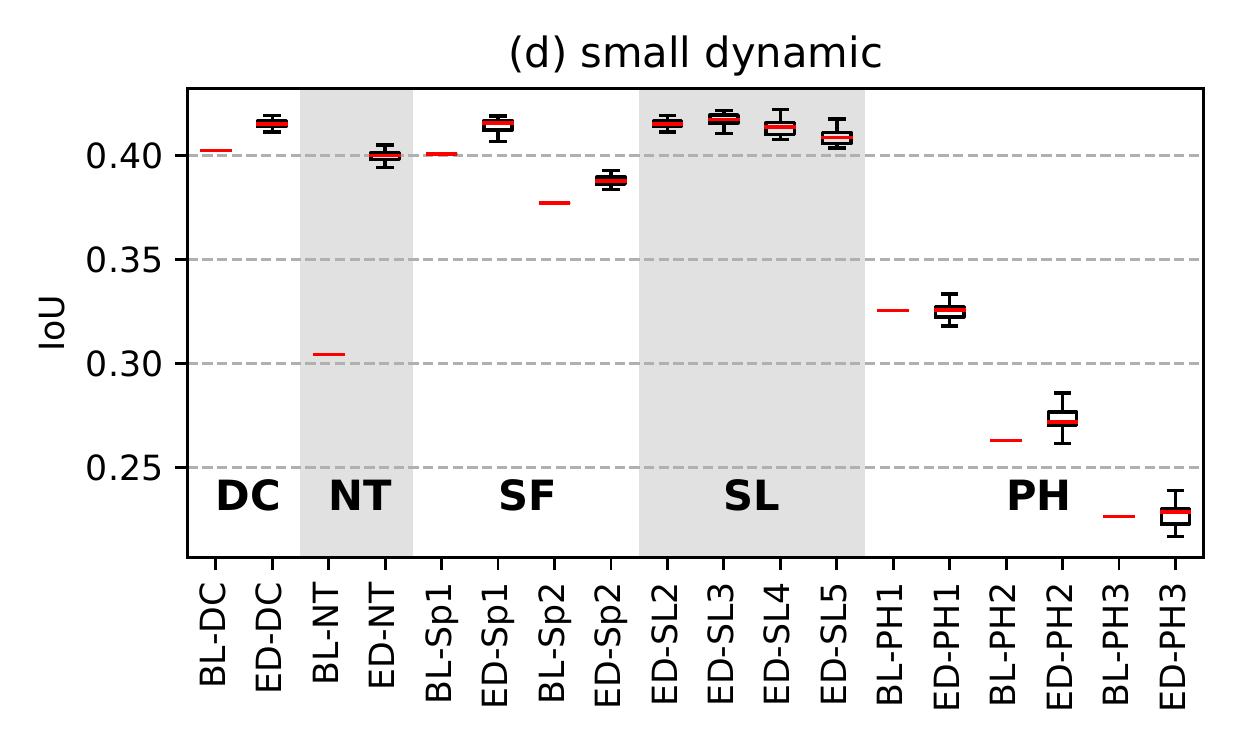}}
  \end{minipage}
  \caption{Category-wise mean IoU reported on the validation set with respect to different framework configurations. Further details on the experiments (columns within the plots) are provided in the paragraphs associated with the corresponding references (DC, NT, SF,...).}
  \label{fig_results}
\end{figure*}

In this section, the behavior of the proposed framework is studied. For this purpose, we have designed several experiments, each analyzing certain aspects of our architecture. For all experiments, the results were analyzed according to categories that contain similarly behaving semantic classes: static (unknown, building, road, sidewalk, vegetation), vehicles (car, bus, truck, train), small static (pole, traffic light, traffic sign) and small dynamic (person, bicycle). Note that bus, truck, and train as well as pole, traffic light, and traffic sign were already combined into one class in the dataset. The class-wise mean intersection over union (IoU) for each of those categories is plotted in Fig.~\ref{fig_results}. The experiment labels (DC, NT, SF,...) are associated with the following paragraphs.

\begin{enumerate}
  \itemsep0em
  \item [DC] \textit{Default config:} We compared our framework utilizing a single sensor (ED-DC) with a simple baseline (BL-DC), which transforms the last sensor frame into the target time.
  \item [NT] \textit{No explicit translation:} We trained the ED to estimate and apply the translational ego motion $q_{t_n}^{\tau}$ without additional external sensor input or the explicit translation step $\mathcal{T}_2$ (ED-NT).
  \item [SF] \textit{Sensor fusion:} We analyze the performance of the framework to fuse grids from multiple simulated cameras (ED-SplX).
  \item [SL] \textit{Sequence length:} We varied the sequence length to study its relation with the prediction performance (ED-SLX).
  \item [PH] \textit{Prediction horizon:} We evaluated how far the network is able to predict into the future (ED-PHX).
\end{enumerate}

\textbf{Baselines:}
To assess the performance of the proposed semantic grid prediction, we designed several baselines that leave out the ED. The simplest baseline BL-NT just replicates the last input grid as prediction $\hat{g}(\tau) = g(t_n)$, while the baseline BL-DC transforms the last input frame using the vehicle's egomotion into the time of the prediction $\hat{g}(\tau) = \mathcal{T}_2(g(t_n), q_{t_n}^\tau)$. To have baselines that consider the difficulties of the sensor split (Sp), we also designed BL-Sp1 and BL-Sp2, which overlay the semantic grid of the lower sensor with the grid of the upper sensor.

\floatsetup[figure]{style=plain,subcapbesideposition=center}
\begin{figure*}
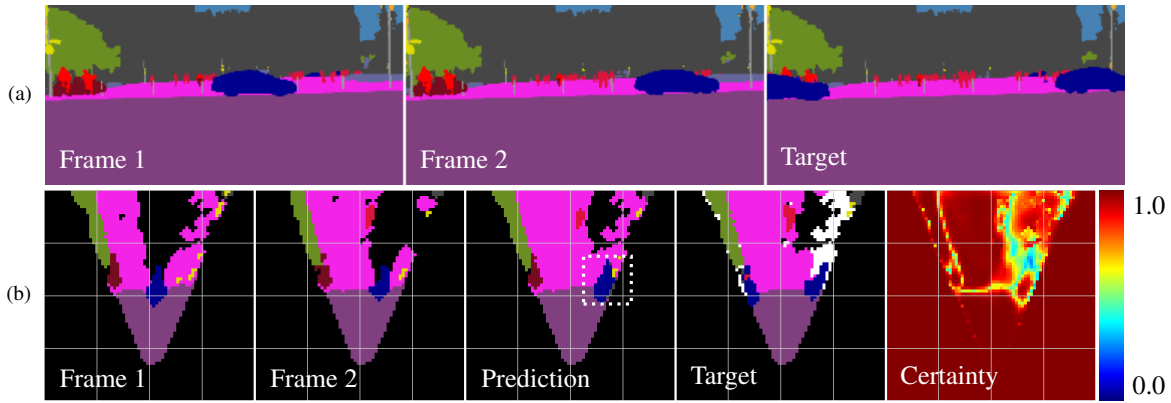

\vspace{0.3cm}
\centering
\sidesubfloat[]{\input{images/figure_sem_5.tex}}\\
\sidesubfloat[]{\input{images/figure_prediction_5.tex}\label{fig_example_1_2}}
\caption{(best viewed in color) Example prediction sequence using ED-DC. (a) shows the sequence of semantic segmentations while (b) presents the associated semantic grids. The framework is fed with frame 1 and 2. The prediction is compared with the target frame. The maximum softmax activation of the prediction is visualized on the right. In this example, a car (blue) is moving to the right. Black areas represent unknown parts of the grid and the white color in the target frame visualizes the loss mask $M$. Note that the left car in the target frame cannot be predicted as it was not visible in frame 1 and 2.}
\label{fig_example_1}
\end{figure*}

\begin{figure*}
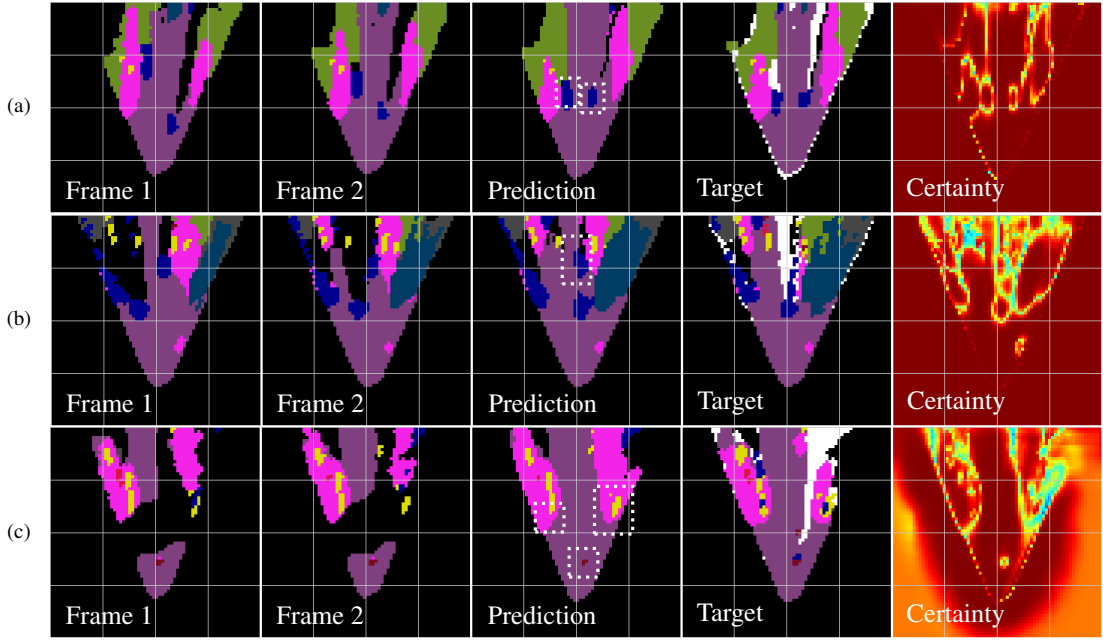

\vspace{0.3cm}
\centering
\sidesubfloat[]{
  \input{images/figure_prediction_6.tex}
  \label{fig_example_overtake}
}\\
\sidesubfloat[]{
  \input{images/figure_prediction_8.tex}
  \label{fig_example_occluded_car}
}\\
\sidesubfloat[]{
  \input{images/figure_prediction_7.tex}
  \label{fig_example_biker}
}
\caption{(best viewed in color) Series of sequences and their associated predictions. (a) The left car (blue) is approaching while the right one drives away (ED-DC). (b) The ED is able to remember the waiting car, which is occluded in the white area of the target frame (ED-DC) (c) A bicycle (maroon) is moving to the right (ED-Sp2). In contrast to the other examples, it uses the Split 2 dataset. Frame 1 and frame 2 are visualized as combination of both sensors. Moreover, the curvature of the left sidewalk towards the blind spot is recognized and correctly completed. Also, the CNN concludes that there is sidewalk around the pole on the right side.}
\label{fig_example_2}
\end{figure*}

\textbf{Default config (DC):}
We compared our framework utilizing a single sensor (ED-DC) with the baseline (BL-DC).
As default ED configuration, we have used $n=2$, $d=3$, and $f=64$, as this configuration provides a good trade-off between performance and runtime%
\ifdefined\ARXIV
~(see supplementary material).
\else
.
\fi
In Fig.~\ref{fig_results_static} and \ref{fig_results_vehicles} column DC, it can be seen that ED-DC significantly outperforms its baseline BL-DC for large static objects and vehicles. On the one hand, the higher performance for large static objects demonstrates the ED's ability to combine the information of the input frames and predict an improved combined grid, which also may contain information about newly occluded areas (see white area of target frame in Fig.~\ref{fig_example_occluded_car}). On the other hand, the high performance for vehicles indicates that the ED is able to predict the motion of dynamic objects and use it for the prediction (see Fig.~\ref{fig_example_1_2}).

Even though classes covering a small area are supposedly quite challenging for ED architectures, our framework is able to outperform the baseline for small dynamic (Fig.~\ref{fig_results_small_dynamic}) and small static (Fig.~\ref{fig_results_small_static}) objects (compare ED-DC with BL-DC). However, the absolute IoU is relatively low for both ED-DC and BL-DC. During qualitative analysis of the data, we found that the effect may be due to temporal noise caused by the alignment of RGB and depth image as well as the grid discretization. Small objects sometimes vanish and appear again making it difficult to predict them. This effect can be mitigated by excluding segmentation borders and regions with a minimum depth difference as well as by increasing the resolution of the semantic grid.

To underline the abilities of the ED CNN as well as the straightforward human-interpretability of the semantic grid, Fig.~\ref{fig:catchy_example}, \ref{fig_example_1}, and \ref{fig_example_2} visualize selected predictions from the validation dataset. In Fig.~\ref{fig_example_1}, a car (blue) is moving perpendicular to the camera. The ED is provided with frame~1 and frame 2. It can be seen, that it translates the car according to its velocity to the right, which is quite accurate compared with the target frame. On the right side of Fig.~\ref{fig_example_1_2}, the maximum softmax activation of the ED is visualized. It can be interpreted as certainty of the CNN's predictions. While the network is certain (red) about already seen static objects such as roads (purple), the maximum softmax activation indicates its uncertainty (green) in the area behind the car and at the borders between classes.
The ED is even able to differentiate between instances of the same class and translate them according to their different velocities. This can be seen in Fig.~\ref{fig_example_overtake}, in which the left car (blue) is approaching while the right one is driving away.
Moreover, the ED can correctly predict areas that would have been occluded in the target frame while they were visible in the frames before (see waiting car in Fig \ref{fig_example_occluded_car} and compare with white area $M$ in target frame).
While the network is able to predict correct classes in most of $M$, it
often fails at the bottom of the grid (see Fig.~\ref{fig:catchy_example}) as there is never ground truth information available during the training process. For a deployable system, we would remove this region from $M$.
\ifdefined\ARXIV
Further examples are given in the supplementary material.
\fi

\textbf{No explicit translation (NT):}
To study the ED's ability to estimate the translational egomotion and the motion of dynamic objects simultaneously and superimpose both, we disabled the explicit translation $\mathcal{T}_2$ of our framework. However, the ED is still provided with the orientationally aligned semantic grids $g(t_i)$. The framework with disabled translation (ED-NT) reaches the performance of the framework with explicit transformation (ED-DC) for large static areas and vehicles (see Fig.~\ref{fig_results_static} and \ref{fig_results_vehicles} column DC and NT). Hence, the network is able to solve both tasks in general. However, it struggles with small objects (see Fig.~\ref{fig_results_small_static} and \ref{fig_results_vehicles}), probably, as a more precise estimate of the egomotion is needed to predict them correctly due to their small size.

\textbf{Sensor fusion (SF):}
In this experiment, the ability of the framework to fuse grids from multiple simulated cameras is evaluated. When using the Split 1 dataset, the framework (ED-Sp1) achieves as good results as without a camera split (ED-DC), as can be seen in Fig.~\ref{fig_results} in the corresponding columns. This shows, that the ED is able to fuse multiple sensors and solve ambiguities. However, a CNN trained on Split 2 (ED-Sp2) performs worse than the other configurations so far, as there is information missing in the input data due to the margins left and right of the sensor and the blind area between both sensors (see Fig.~\ref{fig_extended_split} and \ref{fig_example_biker}). This fact can also clearly be seen in the baseline of Split 2 (see Fig.~\ref{fig_results} BL-Sp2). Even though the missing information decreases the absolute performance, the ED is still able to maintain the difference to its associated baseline compared to other models (see BL-DC/ED-DC and BL-Sp2/ED-Sp2 in Fig.~\ref{fig_results}). For static objects, the CNN is even able to increase the difference as it is able to make assumptions about the environment. For instance, ED-Sp2 concludes that there is sidewalk around the pole on the right side of the grid in Fig.~\ref{fig_example_biker} even though it has never seen the sidewalk before. Note that during training, the semantic grid generated of the whole camera image, instead of the limited image sections for the virtual cameras, was provided as target frame.

\textbf{Sequence length (SL):}
To analyze the influence of the sequence length, we varied $n$.
It can be seen in Fig.~\ref{fig_results} column SL that one additional input frame (ED-SL3) improves the performance of all categories in comparison with ED-SL2 as it probably allows better denoising and speed estimation.
However, a longer sequence length also decreases the number of sequences that can be sampled from the whole training dataset, leading to a drop of performance for ED-SL4/5.

\textbf{Prediction horizon (PH):}
In the last experiment, we have analyzed, how far our framework is able to predict into the future. We have maintained the same step size of about $300~\text{ms}$ between the input frames and evaluated target frames one, two, and three steps after the last input frame (ED-PHX). As expected, the performance of the ED drops with increasing prediction horizon (see Fig.~\ref{fig_results} column PH). Still, the CNN is able to outperform its associated baselines (BL-PHX) for large and medium objects. However, using a long prediction horizon, the ED struggles with small objects, which is probably caused by the increased sensitivity towards spatial mismatches.

%%%%%%%%% DISCUSSION + CONCLUSION
\section{Discussion}\label{sec:conclusion}

Grid-based representations are potentially beneficial in applications where very fast reaction times are required and where the duration to compute and update the ER should be independent of the number of objects.
However, one limitation of grid-based representations is their linear scalability with respect to the number of semantic features.
In the context of autonomous agents, this is especially true for objects carrying a high variety of semantic information, such as road signs. However, the grid
could provide an attention mechanism to identify relevant
areas for further processing (e.g. road sign recognition).

In contrast to other representations, which are not interpretably modularized, our approach provides direct interfaces to verify and test the trained encoder-decoder fusion architecture.
Normally, for camera or laser input signals, it is hard to synthesize new, especially critical, scenarios.
Whereas for our approach, it is comparably easy to generate such sequences of semantic input and according output grid representations.
Further, the interpretable interface functions as a human-readable monitor which can for example be used for debugging or determining corner cases.

Even though we have concentrated on semantic features in
this work, the semantic grid can be extended to support a variety
of other information. As the semantic grid already provides
an abstract, scale-invariant, and low resolution representation
of the environment, down-stream algorithms can
easily extract further information such as correspondence
information or object tracks. If more detailed information
is necessary, additional semantic maps encoding local features
provided by preprocessing steps can be added to the
semantic grid (e.g. local velocities, pedestrian pose, instance
segmentation, or uncertainty).

Other interesting subjects for future work include: multi-scale or dynamic-resolution grid representations, multi-modal sensor inputs (e.g., LiDAR or radar), sensor signals with different dynamics (e.g., frame rates, offsets), and alternative ED architectures for fusion.

In this work, we omitted the estimation of the actual system-inherent latency.
The latency depends on the used hardware and other choices, such as the architecture to extract the semantic information from sensors or the chosen encoder-decoder architecture.
Possible solutions are estimating the delay on the final real-world system and using this estimate for the temporal synchronization of the grid representation before fusion or integrating an additional adaptive component that estimates the current latency.

The presented architecture should be thought of as part of a larger system.
For example, the provided environment representation could be combined with a recurrent representation to store a mid-term ER, a SLAM approach for global mapping, or with a decision making module. In this context, the semantic grid can act as interpretable interface, which can be used for debugging or determining corner cases. Moreover, this interface simplifies synthesizing artificial training data for new, especially critical, scenarios in comparison to the simulation of raw sensor signals.

\section{Conclusion}\label{sec:realconclusion}
We presented and evaluated a concept for generating an environment representation supporting multi-camera fusion on autonomous systems.
We designed the proposed architecture as grid-based to be independent of the number of objects, egocentric to support sensor fusion, interpretable to modularize the system and enhance human accessibility, and finally predictive to compensate for system-inherent latencies.
The architecture was evaluated on the real-world Cityscapes dataset. We demonstrated its superiority to several model-based baselines, its capability to model independent motion of multiple objects, and to fuse ambiguous and incomplete sensor signals.
We think that the proposed architecture and design ideas can further be used as a flexible part of a larger framework to control autonomous systems.

{\small
\bibliographystyle{ieee}
\bibliography{egbib}
}

\ifdefined\ARXIV
	\clearpage
	\section*{Supplementary Material}

% Reset counters and names.
\renewcommand{\thetable}{A\arabic{table}}
\renewcommand{\thefigure}{A\arabic{figure}}
\renewcommand{\thesection}{A\arabic{section}}
\setcounter{table}{0}
\setcounter{figure}{0}
\setcounter{section}{0}

\vspace{1.0em}%
\textbf{Remarks on the semantic grid generation process:}
The used Cityscapes dataset~\cite{cordts2016cityscapes} contains one frame with semantic segmentation groundtruth per sequence. These labeled frames are sufficient to train a semantic segmentation CNN that is able to label the remainder of the sequence frames. For this task, we have used DeepLab~v3~\cite{chen2017rethinking} with the model variant xception 71, the output stride \(8\) and a decoder output stride of \(4\), which achives \(80.31\)\% mean IoU on the test set. Trained weights were provided by the Tensorflow DeepLab GitHub repository\footnote{\url{http://download.tensorflow.org/models/deeplab_cityscapes_xception71_trainvalfine_2018_09_08.tar.gz}}.

For the datasets Split 1 and Split 2, the semantic segmentations are cropped to simulate several sensors each perceiving a specific section of the image. The source semantic segmentation has the resolution \(512 \times 256\) pixels. Split 1 separates the image at $y = 130$, while Split 2 adds a left and right margin of \(40\) pixels and generates the bottom image for $y < 125$ and the top image with $y \geq 145$ leaving a blind spot between both simulated sensors.

To transform the semantic segmentation to the semantic grid, we used the provided preprocessed disparity and camera calibration by Cityscapes to generate a semantic point cloud. In our case the intrinsic camera matrix $K$ and the extrinsic parameters $R_{C\rightarrow A}$ and $t_{C\rightarrow A}$ used in Eq. \ref{eq:camera} can be calculated as follows\footnote{\url{https://github.com/mcordts/cityscapesScripts/blob/master/docs/csCalibration.pdf}}
\begin{equation}
  K =
  \begin{bmatrix}
    f_x & 0 & u_0 & 0\\
    0 & f_y & v_0 & 0\\
    0 & 0 & 1 & 0\\
  \end{bmatrix}
\end{equation}
with $f_x$, $f_y$, $u_0$, $v_0$ representing the focal length and principal point in terms of pixels,
\begin{equation}
  R_{C\rightarrow A} = 
  \begin{bmatrix}
  c_y c_p & c_y s_p s_r - s_y c_r & c_y s_p c_r + s_y s_r \\
  s_y c_p & s_y  s_p s_r + c_y c_r & s_y s_p c_r - c_y s_r\\
  -s_p & c_p s_r & c_p c_r\\
  \end{bmatrix}
\end{equation}
where $s_y$, $s_p$, $s_r$, $c_y$, $c_p$, $c_r$ are the sine and cosine of the extensic yaw, pitch, and roll, and
\begin{equation}
  t_{C\rightarrow A} = (x_{extrinsic}, y_{extrinsic}, z_{extrinsic})^T
\end{equation}
describing the camera to agent translation.

The point cloud is projected onto the ground plane and rotated by the agent's orientation, which is obtained from the ego motion provided by Cityscapes. If two points are assigned to the same grid cell, the class with higher priority with the ascending order static, small static, vehicles, and small dynamic is used. In that way, we ensure that no important information is occluded by another class. Even though by architecture the semantic grid fed into the ED CNN could encode the probabilites of different classes at each grid cell, we have just used one-hot encoded semantic grids, which only represent one class at each grid cell. Class-wise, several morphological operations (dilate, erode, erode, dilate) are applied on the grid to filter noise and close sparsely sampled areas at long range. For large static objects the kernel sizes 3,2,4,4 and for other classes 1,1,2,2 are used.

\vspace{1.0em}%
\textbf{Details on the encoder-decoder CNN:}
The used encoder-decoder (ED) consists of a CNN encoder and a CNN decoder (see Figure \ref{fig_encoder_decoder}). The encoder is built of $d$ blocks. Each block contains \(2\) Convolutional Layers (including batch normalization and ReLu activation) and one max pooling layer. The convolutional kernel size is \(3 \times 3\) and the pooling kernel size \(2 \times 2\). In that way, the spatial resolution in latent space is reduced to \(32 \times 32\) pixels after the third block. Each block doubles the number of feature maps per Convolutional Layer starting with $f$ features. The last block additionally has a dropout layer with a dropout rate of \(0.5\).

The Decoder consists of $d-1$ blocks. Each block upsamples its input and applies a \(2 \times 2\) convolution. The upsampled output is combined with the output of the encoder block with the same size through skip connections and fed into two \(3 \times 3\) Convolutional Layers (including batch normalization and linear activation). The last block has two additional Convolutional Layers reducing the number of feature maps until it is equal to the number of classes. Finally, a softmax is applied.

% \begin{figure*}
% \centering
% \includegraphics[width=7.0in]{figures/EncoderDecoder2.pdf}
% \caption{(best viewed in color) encoder-decoder CNN (ED). The ED combines synchronized semantic grids into one prediction including dynamic object transformation, temporal filtering, and multi-sensor fusion.}
% \label{fig_encoder_decoder}
% \end{figure*}

\vspace{1.0em}%
\textbf{Training configuration:}
For the training of the ED, we used RMSprop with the learning rate $1 \cdot 10^{-3}$ until epoch \(35\) and then switched to $1 \cdot 10^{-4}$. In total, each ED was trained \(40\) epochs with the batch size \(32\) and \(185\) randomly sampled sequences per epoch. For each experimental configuration, the training was repeated \(20\) times. The results were reported on the validations of all iterations. The experiments were run on a Nvidia Pascal Titan.

\vspace{1.0em}%
\textbf{Classwise evaluation:}
Even though, we have combined similar classes in our result analysis to achieve a representive description (see Figure \ref{fig_results}), there are some variances between classes in one category. Therefore, there is a complete overview about the mean IoUs of each class in Table \ref{tab_classresults}.

\begin{table*}
\caption{Mean IoU for each class with respect to the used experimental configuration.}
\label{tab_classresults}
\begin{center}
\begin{tabular}{| c | c | c | c | c | c | c | c | c | c | c |}
\hline
experiment & bicycle & building & car & large vehicle & person & pole/sign & road & sidewalk & unkown & vegetation\\
\hline
color & maroon & gray & blue & turquois & red & yellow & purple & pink & black & green \\
\hline
\hline
BL-DC & 0.41 & 0.59 & 0.52 & 0.55 & \textbf{0.40} & 0.32 & 0.80 & 0.63 & 0.99 & 0.62\\
\hline
ED-DC & \textbf{0.43} & 0.68 & \textbf{0.61} & \textbf{0.64} & \textbf{0.40} & \textbf{0.33} & \textbf{0.86} & \textbf{0.68} & \textbf{1.00} & \textbf{0.72}\\
\hline
BL-NT & 0.30 & 0.57 & 0.46 & 0.52 & 0.30 & 0.21 & 0.80 & 0.58 & 0.99 & 0.61\\
\hline
ED-NT & 0.41 & 0.68 & 0.60 & \textbf{0.64} & 0.39 & 0.31 & \textbf{0.86} & \textbf{0.68} & \textbf{1.00} & 0.71\\
\hline
BL-Sp1 & 0.40 & 0.57 & 0.51 & 0.55 & \textbf{0.40} & 0.32 & 0.78 & 0.58 & 0.99 & 0.58\\
\hline
ED-Sp1 & \textbf{0.43} & 0.68 & \textbf{0.61} & 0.63 & \textbf{0.40} & \textbf{0.33} & \textbf{0.86} & \textbf{0.68} & \textbf{1.00} & \textbf{0.72}\\
\hline
BL-Sp2 & 0.37 & 0.45 & 0.48 & 0.53 & 0.38 & 0.29 & 0.49 & 0.38 & 0.97 & 0.46\\
\hline
ED-Sp2 & 0.39 & 0.61 & 0.57 & 0.60 & 0.38 & 0.29 & 0.81 & 0.60 & 0.99 & 0.63\\
\hline
ED-SL2 & \textbf{0.43} & 0.68 & \textbf{0.61} & \textbf{0.64} & \textbf{0.40} & \textbf{0.33} & \textbf{0.86} & \textbf{0.68} & \textbf{1.00} & \textbf{0.72}\\
\hline
ED-SL3 & \textbf{0.43} & \textbf{0.69} & \textbf{0.61} & \textbf{0.64} & \textbf{0.40} & \textbf{0.33} & \textbf{0.86} & \textbf{0.68} & \textbf{1.00} & \textbf{0.72}\\
\hline
ED-SL4 & \textbf{0.43} & \textbf{0.69} & \textbf{0.61} & 0.63 & \textbf{0.40} & \textbf{0.33} & \textbf{0.86} & \textbf{0.68} & \textbf{1.00} & \textbf{0.72}\\
\hline
ED-SL5 & 0.42 & \textbf{0.69} & 0.61 & 0.63 & \textbf{0.40} & 0.32 & \textbf{0.86} & \textbf{0.68} & \textbf{1.00} & \textbf{0.72}\\
\hline
BL-PH1 & 0.33 & 0.50 & 0.41 & 0.44 & 0.32 & 0.28 & 0.73 & 0.54 & 0.98 & 0.50\\
\hline
ED-PH1 & 0.33 & 0.61 & 0.51 & 0.52 & 0.32 & 0.27 & 0.82 & 0.61 & 0.99 & 0.65\\
\hline
BL-PH2 & 0.26 & 0.43 & 0.35 & 0.36 & 0.27 & 0.23 & 0.67 & 0.48 & 0.98 & 0.41\\
\hline
ED-PH2 & 0.28 & 0.55 & 0.43 & 0.43 & 0.27 & 0.22 & 0.78 & 0.56 & 0.99 & 0.57\\
\hline
BL-PH3 & 0.22 & 0.37 & 0.30 & 0.30 & 0.23 & 0.21 & 0.63 & 0.42 & 0.97 & 0.34\\
\hline
ED-PH3 & 0.24 & 0.49 & 0.37 & 0.33 & 0.21 & 0.18 & 0.75 & 0.51 & 0.99 & 0.50\\
\hline
\end{tabular}
\end{center}
\end{table*}

\vspace{1.0em}%
\textbf{Comparison encoder-decoder configurations:}
To evaluate the importance of the depth $d$ and the number of features starting with $f$ features in the first block, we have varied both parameters and analysed the prediction performance. The results are shown in Figure \ref{fig_ed_results}. Generally, a deeper network with more features performs better than a shallow network with less features. Unfortunately, a higher depth and more features also correspond to a higher runtime (see Figure \ref{fig_runtimes}). As our architecture shall counter system-inherent latencies, it shouldn't consume too much computation time. Therefore, we have chosen the configuration $d=3$ and $f=64$ as default one.

\begin{figure*}
\centering
\includegraphics[width=5in]{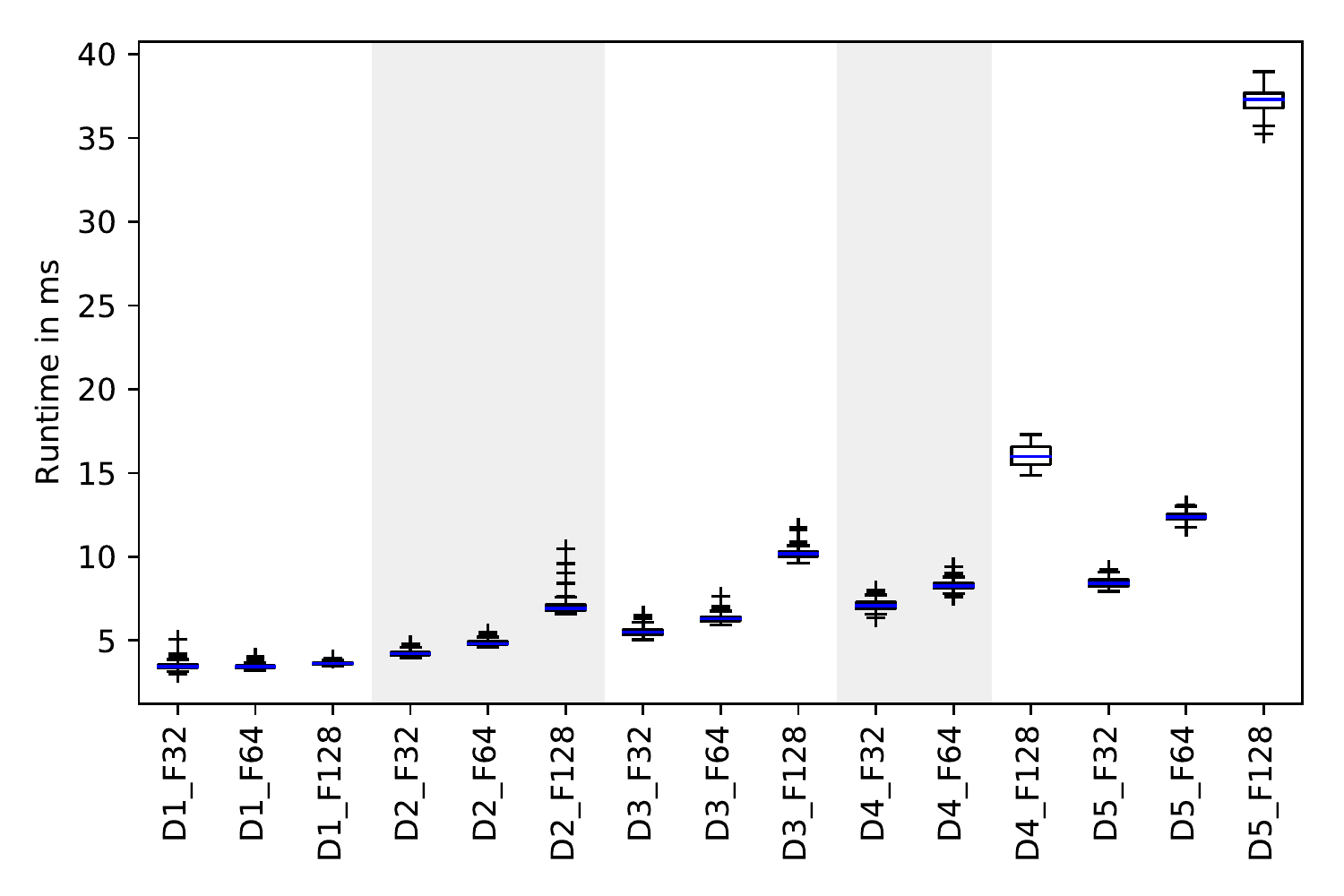}
\caption{Runtimes of different ED configurations. Several network depths $d$ and number of feature maps per convolutional layer starting with $f$ are compared (see Section \ref{sec:experiment_setup} Paragraph Encoder-Decoder CNN). Higher depth and number of features result in an increased runtime. The runtime was evaluated using \(100\) iterations with \(10\) steps each and a batch size of \(1\) on a Nvidia Pascal Titan.}
\label{fig_runtimes}
\end{figure*}

\begin{figure*}
  \centering
  \begin{minipage}{0.49\textwidth}
  \centering
  \subfloat[Mean test IoU for static background (e.g. road and sidewalk).]{%
    \label{fig_results_ed_static}%
    \includegraphics[width=3.2in]{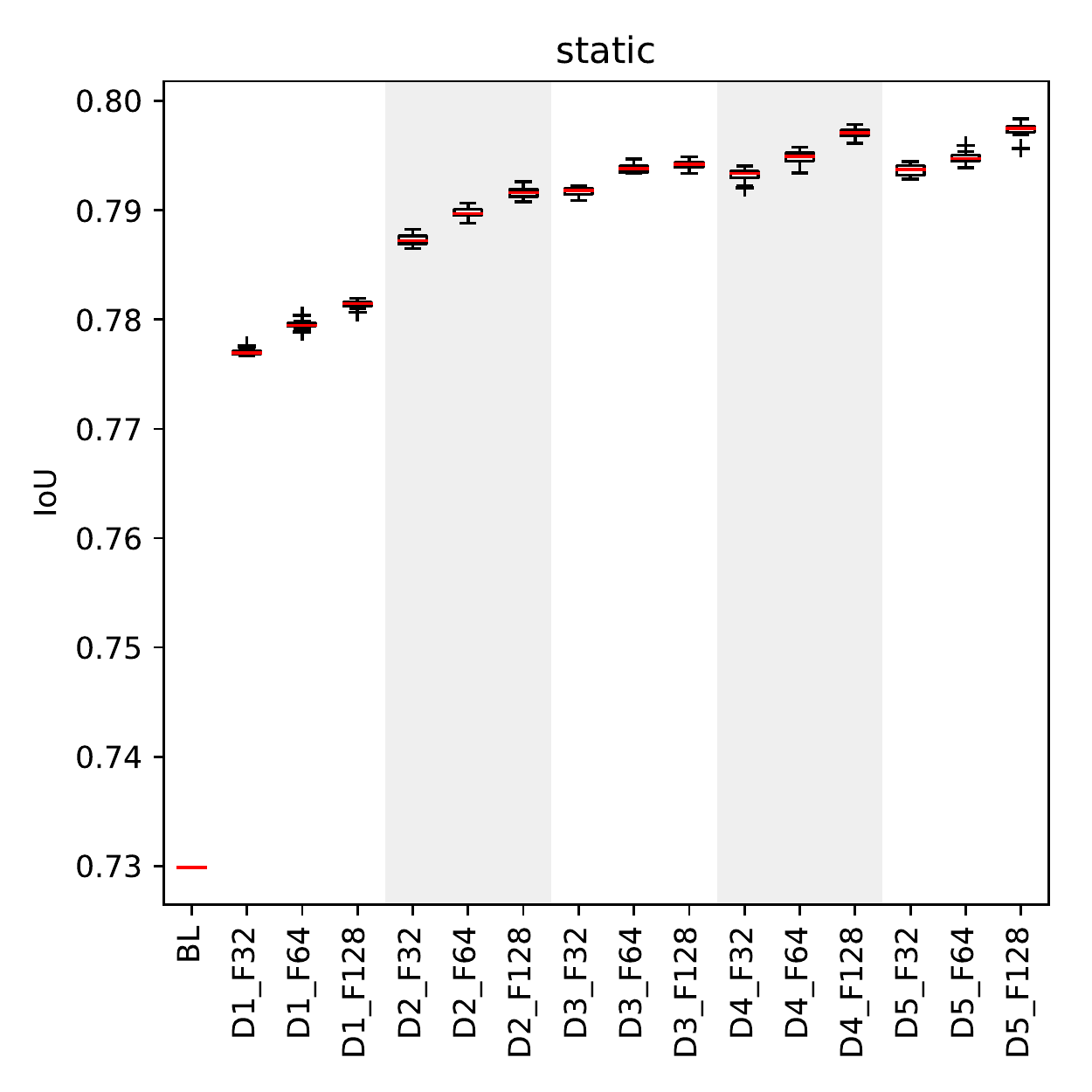}}\\
  \centering
  \subfloat[Mean test IoU for small static objects (pole and traffic sign/light).]{%
    \label{fig_results_ed_small_static}%
    \includegraphics[width=3.2in]{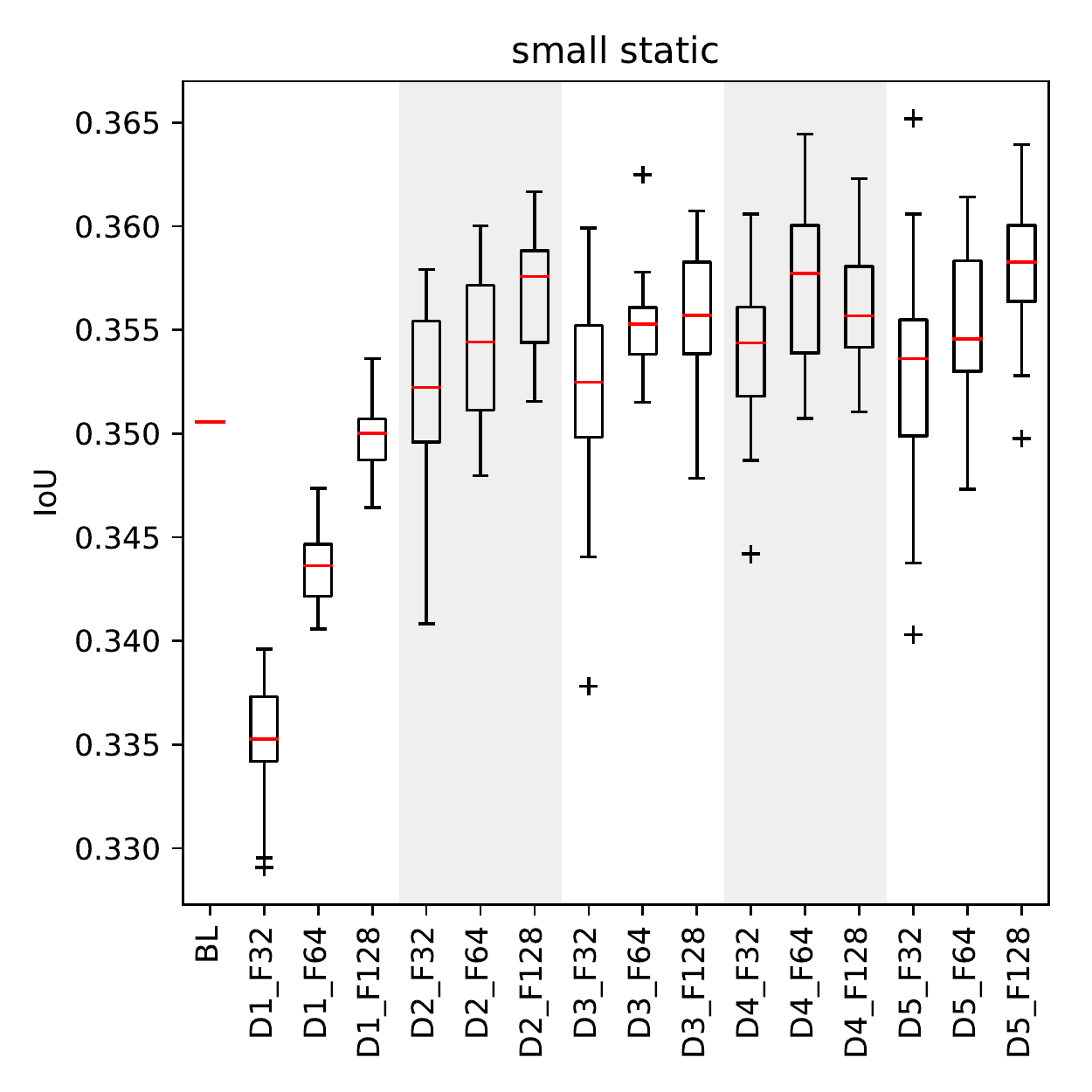}}
  \end{minipage}
  \begin{minipage}{0.49\textwidth}
  \subfloat[Mean test IoU for vehicles (car, bus, truck, and train).]{%
    \label{fig_results_ed_vehicles}%
    \includegraphics[width=3.2in]{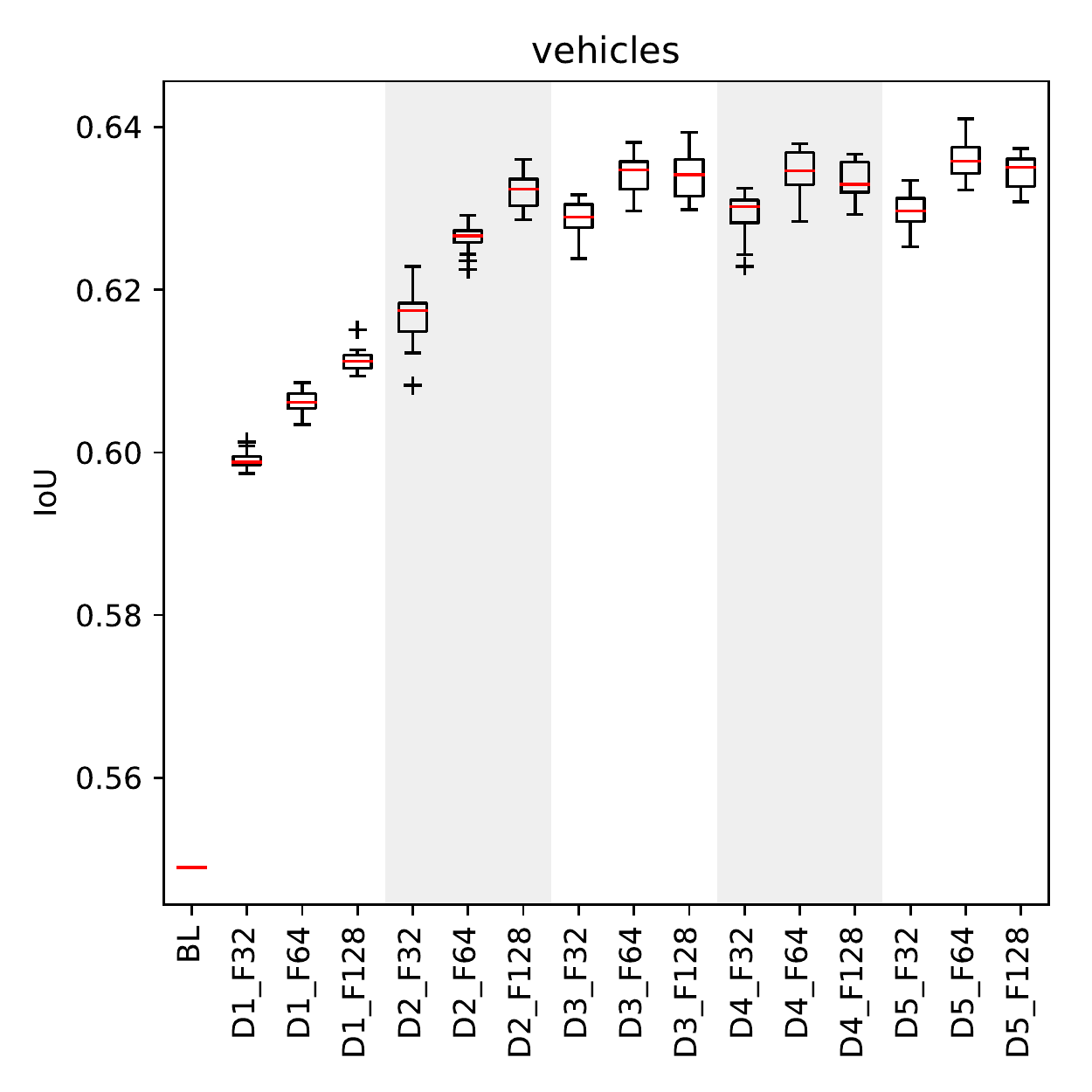}}\\
  \subfloat[Mean test IoU for small dynamic objects (bicycle and person).]{%
    \label{fig_results_ed_small_dynamic}%
    \includegraphics[width=3.2in]{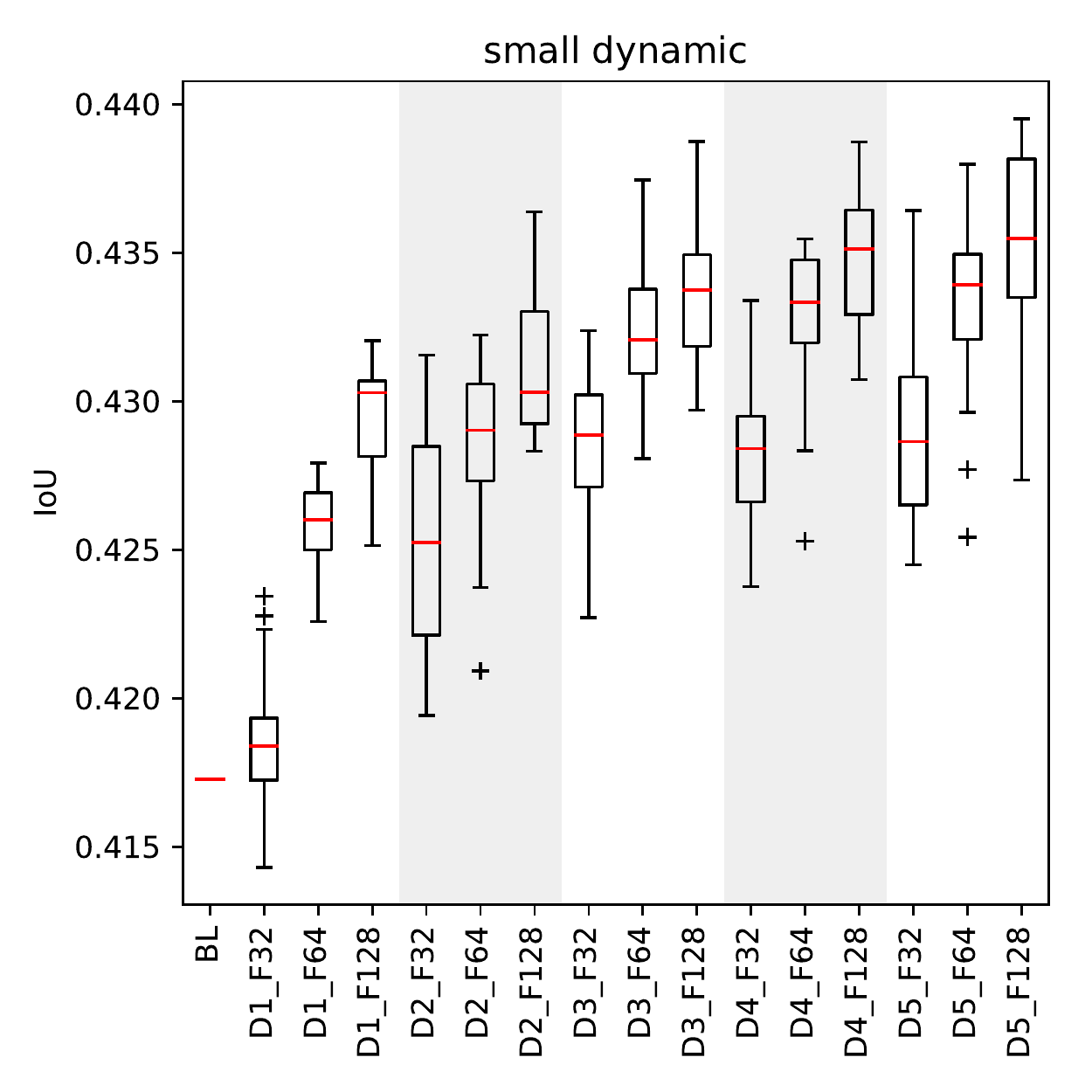}}
  \end{minipage}
  \caption{Comparison of different ED configurations with respect to the mean test IoU. It can be seen, that a higher depth~$d$ and more features~$f$ generally yield higher test accuracies.}
  \label{fig_ed_results}
\end{figure*}

\vspace{1.0em}%
\textbf{Systematically selected examples:}
In order to understand why the ED performs better than the baselines, we have systematically evaluated example predictions on the validation dataset. We have selected sequences, where the IoU of the prediction most outperforms both baselines BL-DC and BL-NT with the metric $\text{min}(\text{IoU}_{\text{ED-DC}} - \text{IoU}_{\text{BL-DC}}, \text{IoU}_{\text{ED-DC}} - \text{IoU}_{\text{BL-NT}})$. In Fig. \ref{fig_example_car} and \ref{fig_example_road} representative examples of the top \(10\) sequences according to that metric are shown for the classes car and road.

\begin{figure*}
\centering
\input{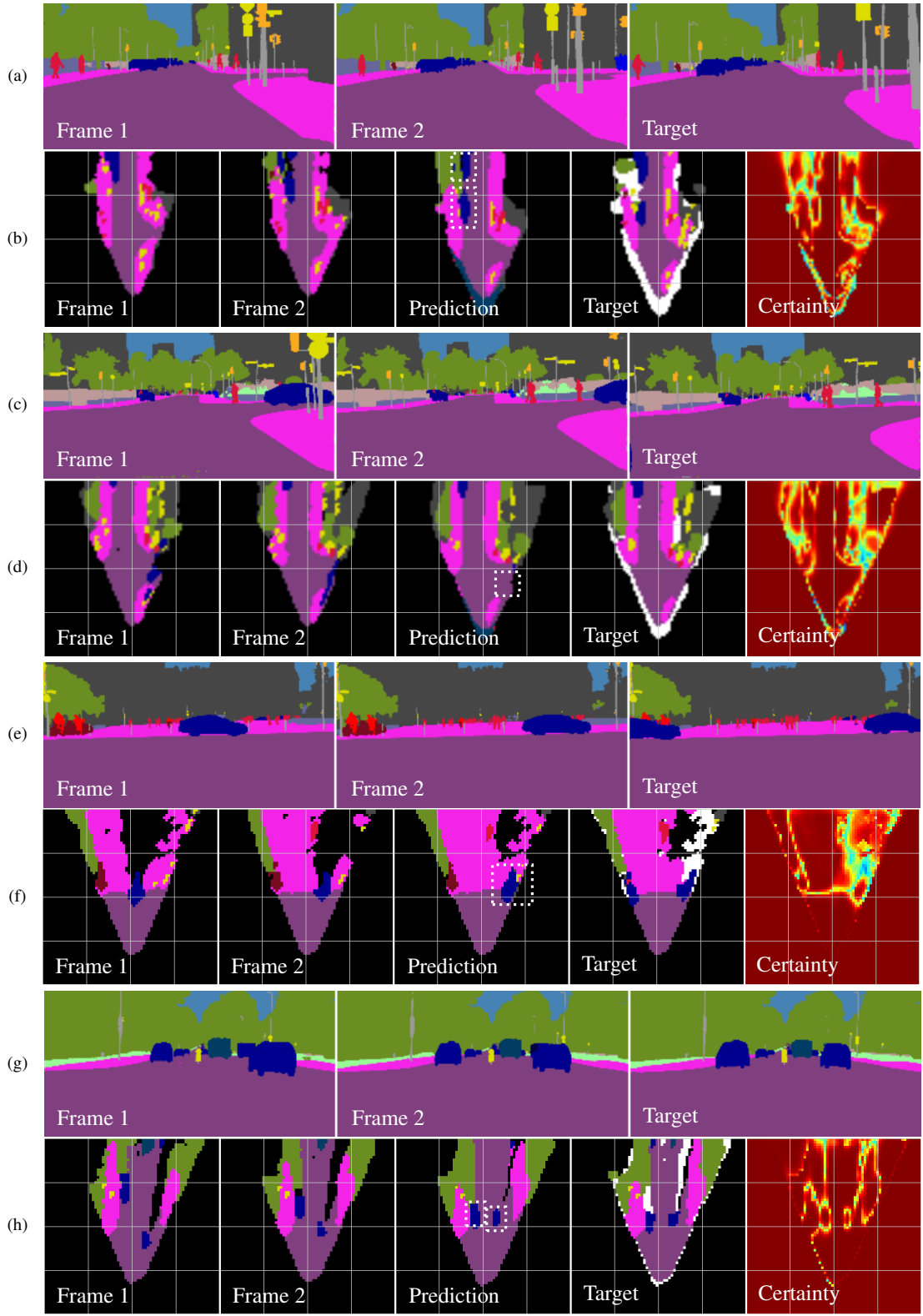}
\caption{Example predictions, where ED-DC most outperforms BL-DC and BL-NT, for the class car (blue) on the validation dataset. (a/b) Two cars are separating from each other. (c/d) The right car is leaving the semantic grid. (e/f) A car is moving perpendicular to the agent. (g/h) The right car is driving in front and the left car is approaching.}
\label{fig_example_car}
\end{figure*}

\begin{figure*}
\centering
\input{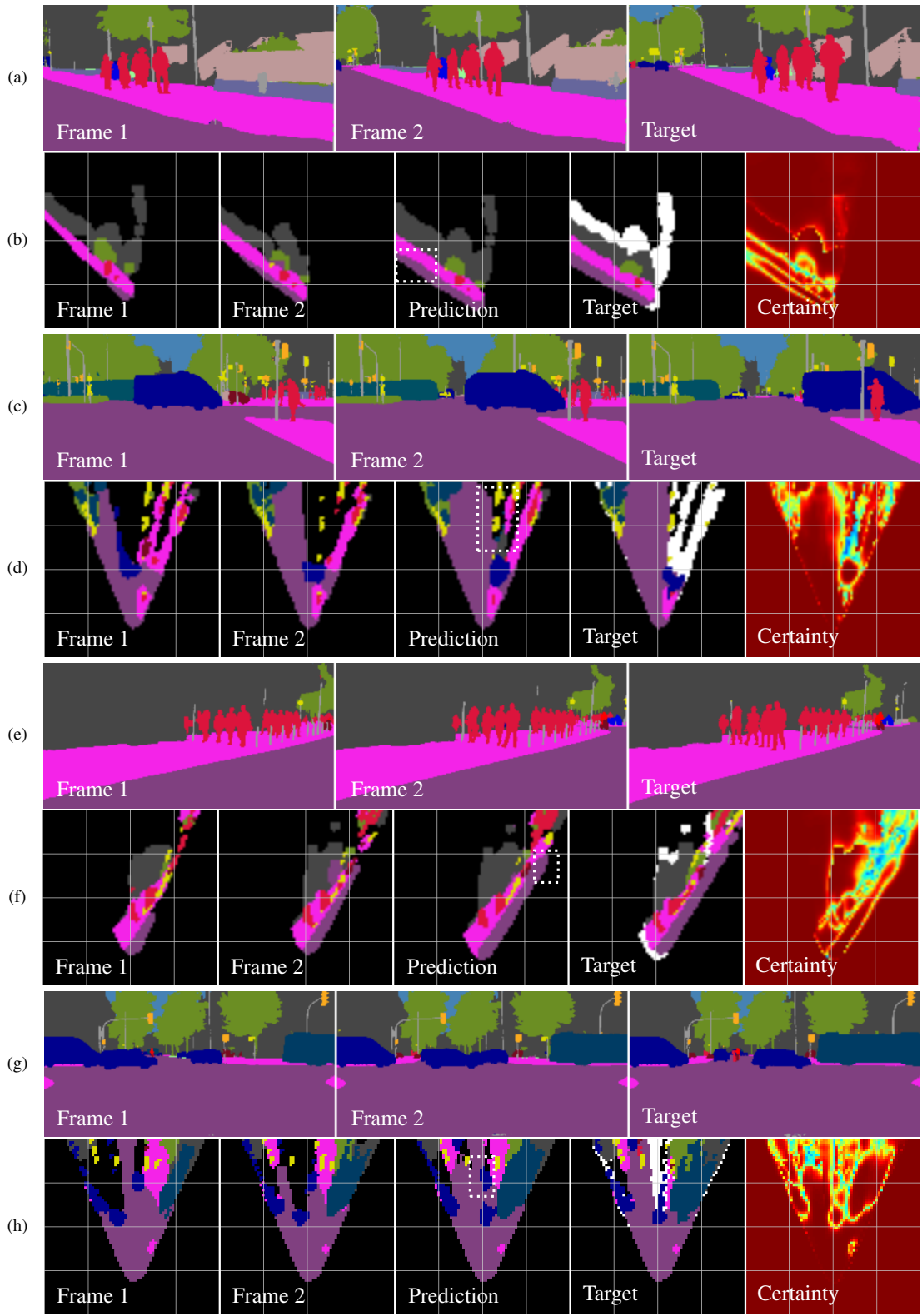}
\caption{Example predictions, where ED-DC most outperforms BL-DC and BL-NT for the class road (purple) on the validation dataset. (a/b) Prediction of road left to the sidewalk (pink). (c/d) Prediction of road behind car (blue). (e/f) Prediction of road right to the sidewalk. (g/h) Prediction of road in the shadow of the cars. Additionally, this example shows that the CNN is able to remember the car in the shadow of the other one.}
\label{fig_example_road}
\end{figure*}

\vspace{1.0em}%
\textbf{Example predictions of ED-Sp2:}
In Figure \ref{fig_example_split2}, example predictions using the architecture ED-Sp2 on the Split 2 dataset with simulated cameras (see Fig. \ref{fig_extended_split}) are shown to underline the ED's ability to deal with ambiguous and incomplete multi-sensor data.

\begin{figure*}
\centering
\input{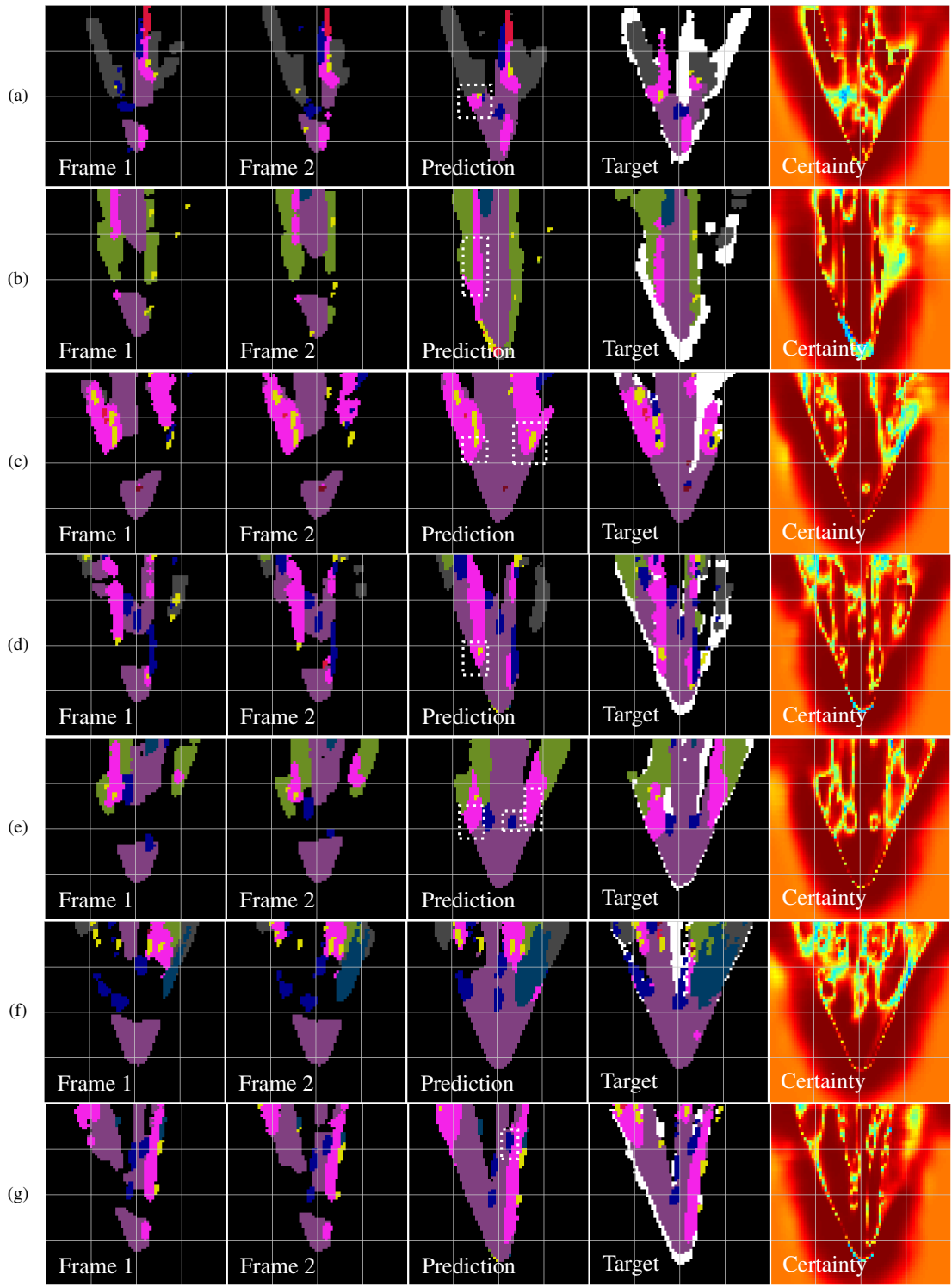}
\caption{Example predictions for the Split 2 dataset (validation). Frame 1 and frame 2 are the combination of both sensors using BL-Sp2 for visualization purpose. ED-Sp2 is able to make mostly correct assumptions about the blind sensor areas. (a) A sidewalk (pink) is predicted between road (purple) and building (gray) on the left side. (b) The sidewalk on the left side becomes connected. (c) The curvature of the left sidewalk towards the blind spot is recognized and correctly completed. Moreover, the CNN concludes that there is sidewalk around the pole on the right side. (d) The sidewalk on the left side is correctly extended. (e) The right car is properly predicted while in frame 2 only a small portion is visible. Moreover, the sidewalk on both sides of the street is also predicted correctly. (f) All cars are tracked correctly. (g) The car turning right is predicted correctly.}
\label{fig_example_split2}
\end{figure*}

\vspace{1.0em}%
\textbf{Example predictions for different prediction horizons:}
In Figure \ref{fig_example_prediction_horizon}, example predictions with different prediction horizons are shown for ED-PH1, ED-PH2, and ED-PH3.

\begin{figure*}
\centering
\input{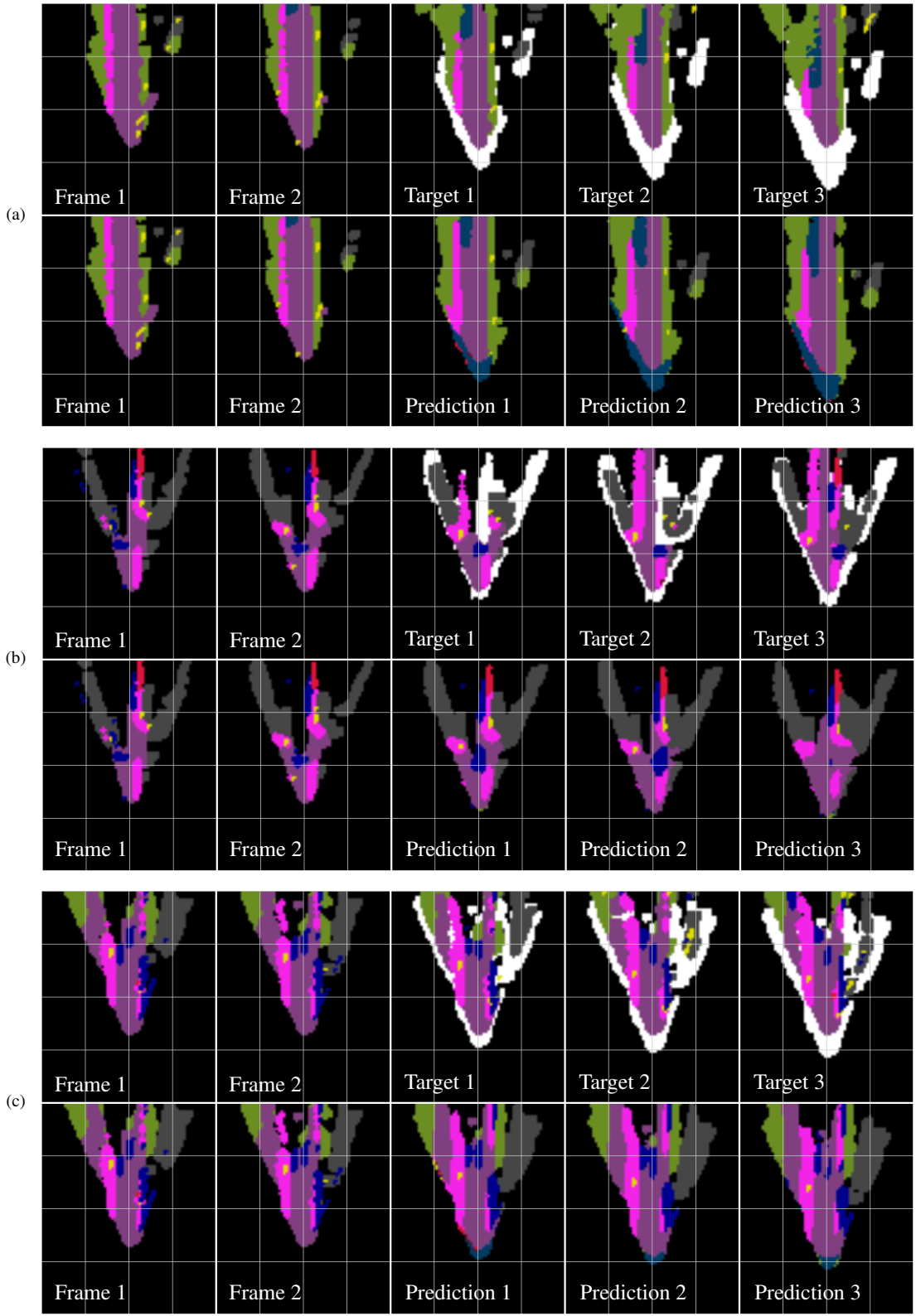}
\caption{Examples for different prediction horizons. In each case, the upper row visualizes both input and the three target frames. The bottom row shows both input frames and the predictions for different prediction horizons associated with the target frames. It can be seen that the ED correctly predicts most static objects. Remarkably in (a), the bus is predicted correctly up to Prediction 3 just given Frame 1 and Frame 2. However, in some cases the ED struggles with precisely estimating the motion of other cars with a larger prediction horizon (see (b) and (c) Prediction 3).}
\label{fig_example_prediction_horizon}
\end{figure*}

\fi

\end{document}